\documentclass{article}

     \PassOptionsToPackage{numbers, compress}{natbib}

     \usepackage[preprint]{neurips_2022}

\usepackage[utf8]{inputenc} 
\usepackage[T1]{fontenc}    
\usepackage{hyperref}       
\usepackage{url}            
\usepackage{booktabs}       
\usepackage{amsfonts}       
\usepackage{nicefrac}       
\usepackage{microtype}      
\usepackage{xcolor}         

\usepackage{multirow}
\usepackage{comment}
\usepackage{amsmath}
\usepackage{mathtools}
\usepackage{adjustbox}
\usepackage{subcaption}
\usepackage[symbol]{footmisc}

\newcommand{\slfrac}[2]{\left.#1\middle/#2\right.}

\title{Cross-Modal Alignment Learning of \\ Vision-Language Conceptual Systems}

\author{%
  Taehyeong Kim$^1$, Hyeonseop Song$^1$, Byoung-Tak Zhang$^{2}$ \\
  $^1$LG Electronics, $^2$Seoul National University \\
  \texttt{\{taehyeong.kim,hyeonseop.song\}@lge.com, btzhang@bi.snu.ac.kr} \\
}

\begin{document}

\maketitle

\begin{abstract}
Human infants learn the names of objects and develop their own conceptual systems without explicit supervision.
In this study, we propose methods for learning aligned vision-language conceptual systems inspired by infants' word learning mechanisms.
The proposed model learns the associations of visual objects and words online and gradually constructs cross-modal relational graph networks.
Additionally, we also propose an aligned cross-modal representation learning method that learns semantic representations of visual objects and words in a self-supervised manner based on the cross-modal relational graph networks.
It allows entities of different modalities with conceptually the same meaning to have similar semantic representation vectors.
We quantitatively and qualitatively evaluate our method, including object-to-word mapping and zero-shot learning tasks, showing that the proposed model significantly outperforms the baselines and that each conceptual system is topologically aligned.
\end{abstract}

\section{Introduction}
\label{sec:intro}

According to studies by psychologists and cognitive scientists, infants could categorize visual objects (hereafter objects) at the age of 3-4 months \cite{quinn2001perceptual}, segment words from fluent speech at the age of 7-8 months \cite{jusczyk1995infants}, and finally learn a language.
How do infants learn a language and build their own conceptual systems \cite{mandler2007origins}?
Although it is not clearly known about these developmental processes, various studies have been conducted on the word learning mechanisms of infants \cite{swingley2009contributions}, which could be a fundamental process for building vision-language conceptual systems.

Infants begin to learn the meaning of words by pairing heard words with visually present objects.
It is not an easy process because the mapping of specific words to multiple objects in a visual scene is ambiguous \cite{quine2013word}.
For example, when their parents say the word ``cup'' when ``spoon'', ``cup'', ``food'', or ``plate'' are placed on the table, infants may not know exactly which visual object it refers to.
Nevertheless, typical infants know the meaning of some common nouns after the age of 6-9 months \cite{bergelson20126}, which is a crucial stage for learning more words and acquiring language \cite{swingley2009contributions}.
Although the mechanisms by which infants learn words and language remain unclear, the best-known mechanism is cross-situational learning (XSL) \cite{he2017word}.
XSL is that even if it is ambiguous whether a word refers to a visual object in a scene, the mapping of words and visual objects becomes more and more accurate if sufficient information is obtained through continuous exposure to various situations.

Meanwhile, computer scientists have also studied methods for learning the mappings between objects and words in an unsupervised or weakly-supervised manner.
The visual grounding (VG) seeks to find the target objects or regions in an image based on a natural language query \cite{yu2018rethinking}.
Various methods \cite{deng2018visual, liu2020learning, xiao2017weakly} to solve the VG task have been studied, but they usually only map specific words or noun phrases and objects when an image and text are given as a pair, and it is not easy to generalize the relationship between objects and words.
Aligned cross-modal representation learning also attempts to map between images and natural language \cite{jia2021scaling,radford2021learning}.
It builds embeddings that similarly represent objects and words with the same meaning in a common semantic space, but do not explicitly represent their complex relationships.

\begin{figure}
\vskip -0.1in
\begin{center}
\centerline{\includegraphics[width=0.85\textwidth]{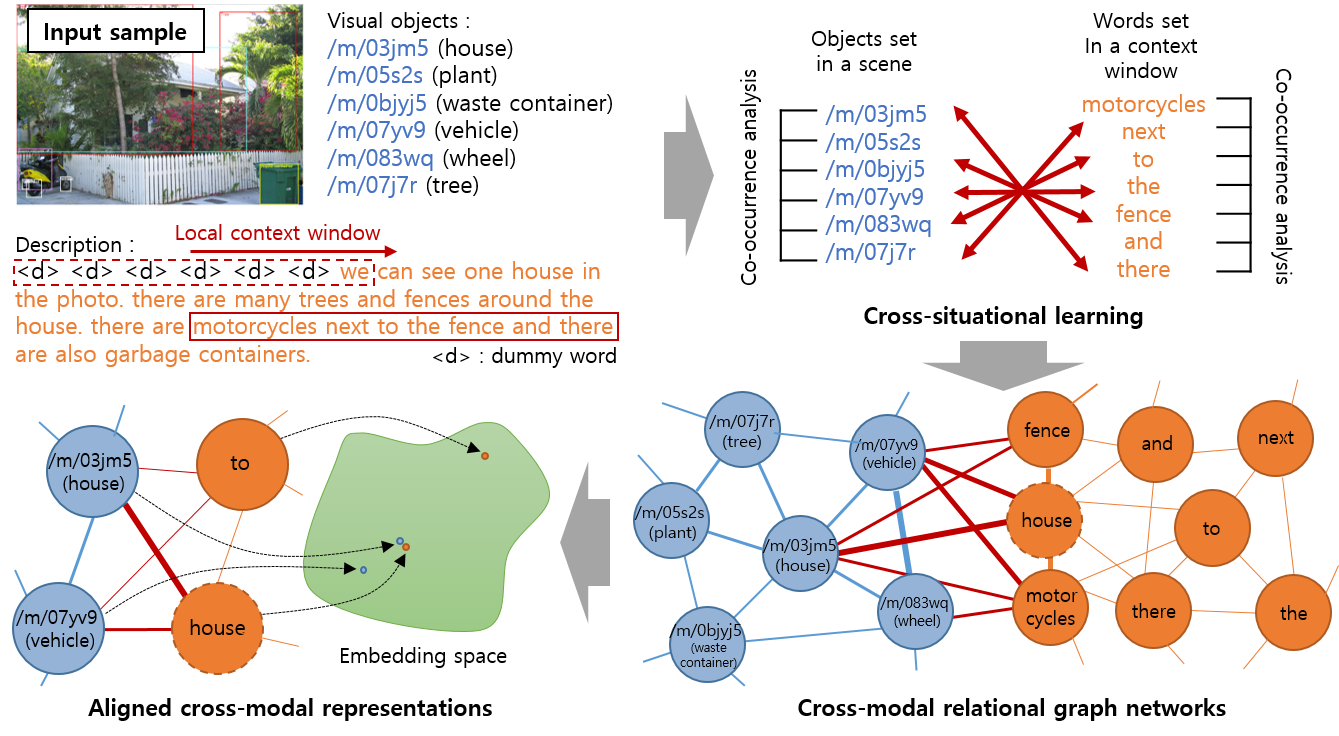}}
\caption{From the input sample, i) our model continuously learns object-object, word-word, and object-word co-occurrence statistics to construct cross-modal relational graph networks. ii) Based on the graph networks, it learns aligned semantic representations so that objects and words with conceptually the same meanings have similar representation vectors.}
\label{fig:overview}
\end{center}
\vskip -0.2in
\end{figure}

From these backgrounds, we propose a continuous learning method of vision-language conceptual systems, inspired by the word learning mechanisms.
We assume that the learner's object categorization and word segmentation abilities have already been developed; thus, we use image-level pseudo labels (or a pre-trained object detector) and text data.
The proposed model builds cross-modal relational graph networks by continuously learning the co-occurrence statistics of objects and words from the image-text dataset based on the distributional hypothesis \cite{harris1954distributional}.
The distributional hypothesis states that linguistic items with similar distributions have similar meanings, which could be applied to objects in visual scenes as well as natural language.
We use simple statistical rules to progressively construct object-relational and word-relational graph networks representing vision and language conceptual systems.
Furthermore, we utilize the object-word co-occurrence statistics to align the two graph networks.
These methods continuously reduce the ambiguity of object and word mapping, and gradually build vision-language conceptual systems (i.e. cross-modal relational graph networks).
We additionally propose a novel method for learning aligned cross-modal representations based on the built relational graph networks.
The proposed method trains the semantic representation vectors of objects and words in a self-supervised manner.
We evaluate our model quantitatively and qualitatively, showing that the graph networks and semantic representations learned by the proposed methods are topologically aligned. An overview of the proposed methods is depicted in Figure \ref{fig:overview}.

Our main contributions can be summarized as follows.
\begin{itemize}
\item We propose a method for progressively building cross-modal relational graph networks by continuously learning object-object, word-word, and object-word co-occurrence statistics.
\item We propose an aligned cross-modal representation learning method that learns semantic representations of objects and words through cross-entropy loss for self-identification of each entity and alignment loss for correspondence between objects and words.
\item Finally, we validate our model with object-to-word mapping accuracy and zero-shot learning tasks. The experimental results show that the proposed methods significantly outperform the baselines and that each conceptual system is topologically aligned.
\end{itemize}

\section{Related work}
\label{sec:related}

\textbf{Cross-situational learning}\quad
%
Various studies \cite{clerkin2019everyday, suanda2014cross, vlach2013memory, yu2007rapid} discuss XSL strategies based on distribution statistics in which words and referents occur simultaneously at multiple moments.
There are works \cite{he2017word, rasanen2015joint} that examine how learners segment words and acquire their meanings through XSL.
\citet{fazly2010probabilistic} and \citet{kadar2015learning} propose computational models that learn word meanings by probabilistic associations between words and semantic elements. 
There are also studies using the XSL mechanism for robot learning.
\citet{taniguchi2017cross} proposes a Bayesian generative model that can form multiple categories based on each sensory channel of a humanoid robot and can associate words with four sensory channels.
\citet{roesler2018probabilistic} introduces a Bayesian learning model for grounding synonymous object and action names using XSL.
Recently, deep neural network models have been proposed to learn semantic knowledge \cite{nikolaus2021evaluating} and representations \cite{khorrami2021can} using images with corresponding descriptions and audio.

\textbf{Cross-modal representation learning}\quad
\citet{castrejon2016learning} introduces a new cross-modal scene dataset, and presents a method for training cross-modal scene representations to have a shared representation that is modal-agnostic.
CLIP \cite{radford2021learning} demonstrates that a simple pre-training task of predicting which caption goes with which image is an efficient and scalable way to learn image representations from large datasets of image-text pairs.
ALIGN \cite{jia2021scaling} consists of a simple dual-encoder architecture and learns to align visual and language representations of image and text pairs using contrastive loss.
OSCAR \cite{li2020oscar} learns cross-modal representations using masked token loss and contrastive loss with image-text and object tags.
The most relevant work to ours is \cite{roads2020learning}, which proposes a way to align conceptual systems in an unsupervised manner using representation vectors trained with the GloVe \cite{pennington2014glove} method.
These methods are difficult to explicitly learn complex relationships between entities. We solve this problem by constructing the cross-modal relational graph networks through a cross-situational learning method.

\section{Cross-modal relational graph networks}
\label{sec:relational}

\subsection{Problem settings and preliminaries}

In this section, we introduce a method to build cross-modal relational graph networks by continuously learning the co-occurrence statistics of objects and words.
To investigate the proposed methods, we use image-level pseudo labels (i.e. categories of objects included in the image) and text descriptions for images in the Open Images Dataset V6 \cite{OpenImages}.
Formally, given an image $I$ containing $M$ objects, we use the set of objects $B={\{b_m\}}_{m=1}^M$, where $b_m$ denotes the pseudo label of the object.
Also, given a text description $S$ corresponding to the image $I$, we use the word sequence $S=[s_n]_{n=1}^N$ of length $N$, where $s_n$ denotes the word tokenized by NLTK \cite{bird2006nltk}.
Our goal is to build cross-modal relational graph networks $G(G^O,G^W,E^X)$ by continuously learning the relationships between objects and words from the input of $B$ and $S$.
Here, $G^O$ and $G^W$ represent weighted relational graph networks for objects and words, respectively, and edge set $E^X$ represents associations between the two graphs.

To this end, we generate multi-modal sequential inputs $[(B,Q_t)]_{t=1}^T$ using local context words $Q_t$ for each paired input $(B,S)$ to simulate the stream input and reflect the local context.
To explain local context words $Q_t$, we first define a padded word sequence $\tilde{S} = [D;S]=[\tilde{s}_t]_{t=1}^T$ filled with dummy words $D$ at the beginning of the sequence $S$. 
The length of the dummy word sequence $D$ is determined as $H-1$ for the size $H$ of the local context window.
Therefore, the length $T$ of the padded word sequence $\tilde{S}$ is determined as $N+H-1$.
The local context $Q_t={\{q_k\}}_{k=1}^K$ refers to set of all words included in the local context window of $\tilde{s}_t$ (i.e. $H$ words after $\tilde{s}_t$ including itself) in the padded word sequence $\tilde{S}$.
Finally, we get multi-modal sequential inputs $[(B,Q_t)]_{t=1}^T={[(\{b_1,b_2,...,b_m\},\{q_1,q_2,...,q_k\})]}_{t=1}^T$ for each input pair $(B,S)$.
In the following subsections, we first describe methods to form the undirected weighted graph networks $G^O$ and $G^W$ from these stream inputs, and then we describe how to build a cross-modal relational graph networks $G(G^O,G^W,E^X)$ with edge set $E^X$ using the proposed XSL method.
Graph networks $G$ is described as an undirected graph, and edge $(v_a,v_b)$ and edge $(v_b,v_a)$ connecting the two nodes $v_a$ and $v_b$ are equivalent.

\subsection{Object and word graph networks}
In natural language, semantically similar words occur together more often than unrelated words.
Based on this assumption, we incrementally build the relational graph networks $G^W(V^W,E^W)$ of words, where $V^W$ is the set of word nodes and $E^W$ is the set of edges between the nodes.
Each word node $v_{w_i}$ has a variable $d_{w_i}$ representing the cumulative number of inputs for the word $w_i$.
For each local context $Q_t=\{q_1,q_2,...,q_k\}$ in the input stream, if a new word $w_{new}$ not included in $V^W$ is observed, a new node $v_{w_{new}}$ is created, and $d_{w_{new}}$ is initialized to 0.
Then $d_{q_k}$ is increased by 1 for every node $v_{q_k}^W$ corresponding to all words $q_k$ in $Q_t$.

Each edge $(v_{w_i},v_{w_j})$ has the number of co-occurrence $c_{w_iw_j}$ of the two words $w_i$ and $w_j$ as a variable.
For each local context $Q_t=\{q_1,q_2,...,q_k\}$ in the input stream, when a new node $v_{w_{new}}$ is created, edges $(v_{w_{new}},v_{w_*})$ connecting that node $v_{w_{new}}$ and all other nodes $v_{w_*}$ are created, and $c_{w_{new}w_*}$ is initialized to 0.
Then, for every word pairs $(q_i,q_j)$ in $Q_t$, the co-occurrence count $c_{q_iq_j}$ is increased by 1.
By these processes, the connection strength (i.e. weight of edge) $e_{w_iw_j}$ between two nodes $v_{w_i}$ and $v_{w_j}$ is defined as Equation \ref{eq:weights}.
The edge weight $e_{w_iw_j}$ has a value of 1 for the two words $w_i$ and $w_j$ that always appear together, and the lower the frequency of their co-occurrence, the closer to 0.
In this way, the distribution and relationship of words are learned, and the undirected weighted graph networks $G^W$ is continuously formed.

Similarly in vision, semantically related objects appear together more often in a scene.
Based on this assumption, we also build the relational graph networks $G^O(V^O,E^O)$ of objects in the same way as the relational graph networks formation of words.
Here, $V^O$ is the set of object nodes and $E^O$ is the set of edges between the nodes.
To form the relational graph networks for objects, we use the set of the objects $B$ in the input stream instead of the local context $Q_t$ of word sequence $S$.
For each object set $B=\{b_1,b_2,...,b_m\}$, if a new object $o_{new}$ not included in $V^O$ is observed, a new node $v_{o_{new}}$ is created, and $d_{o_{new}}$ and $c_{o_{new}b_*}$ are initialized to 0. 
Then $d_{b_m}$ is increased by 1 for every node $v_{b_m}$ corresponding to every object in $B$, and the number of co-occurrences $c_{b_ib_j}$ for edge ($v_{b_i}$,$v_{b_j}$) for every pair of objects $(b_i,b_j)$ in $B$ is also increased by 1.
Finally, using the cumulative input counts $d_{o_i}$ and $d_{o_j}$ of the two object nodes $v_{o_i}$ and $v_{o_j}$ and their co-occurrence count $c_{o_io_j}$, the edge weight $e_{o_io_j}$ of the object nodes is defined as shown in Equation \ref{eq:weights}.
\begin{equation} \label{eq:weights}
e_{w_iw_j} = \nicefrac{c_{w_iw_j}^2}{(d_{w_i} \cdot d_{w_j})}, \qquad
e_{o_io_j} = \nicefrac{c_{o_io_j}^2}{(d_{o_i} \cdot d_{o_j})}
\end{equation}
Following the formation of relational graph networks $G^W$ and $G^O$, the next subsection describes how to form relationships between objects and words with edge set $E^X$.

\subsection{Cross-situational learning}
\label{sec:xsl}
XSL paradigm allows learners to gradually acquire the meaning of words through multiple exposures to the environment.
We implement XSL for incremental learning of the relationships between objects and words using a cross-mapping approach.
For the set $E^X$ of edges connecting all object nodes $v_{o_*}$ and word nodes $v_{w_*}$, each edge $(v_{o_i},v_{w_j})$ has the cumulative number of co-occurrence $c_{o_iw_j}$ as a variable for the object $o_i$ and word $w_j$.
For each multi-modal input $(B,Q_t)=(\{b_1,b_2,...,b_m\},\{q_1,q_2,...,q_k\})$, when a new object node $v_{o_{new}}$ is generated, edges $(v_{o_{new}},v_{w_*})$ connecting $v_{o_{new}}$ and all word nodes $v_{w_*}$ are created and the co-occurrence counts $c_{o_{new}w_*}$ are initialized 0.
Likewise, when a new word node $v_{w_{new}}$ is generated, edges $(v_{o_*},v_{w_{new}})$ connecting $v_{w_{new}}$ and all object nodes $v_{o_*}$ are created and $c_{o_*w_{new}}$ are initialized 0.
Then, for all object-word pairs $(b_i,q_j)$ in $(B,Q_t)$, $c_{b_iq_j}$ is increased by 1.
The connection strength $e_{o_iw_j}$ between the object node $o_i$ and word node $w_j$ is formalized as Equation \ref{eq:cross_weight}.
%
\begin{equation}
\label{eq:cross_weight}
\begin{gathered}
a_{o_i}(w_j)= \slfrac{\frac{c_{o_iw_j}^2}{d_{o_i} \cdot d_{w_j}}}{\sum_{w_u \in V^W}\frac{c_{o_iw_u}^2}{d_{o_i} \cdot d_{w_u}}}, \qquad 
a_{w_j}(o_i)= \slfrac{\frac{c_{o_iw_j}^2}{d_{o_i} \cdot d_{w_j}}}{\sum_{o_u \in V^O}\frac{c_{o_uw_j}^2}{d_{o_u} \cdot d_{w_j}}} \\
e_{o_iw_j}=a_{o_i}(w_j) \cdot a_{w_j}(o_i)
\end{gathered}
\end{equation}
%
In this equation, $a_{o_i}(w_j)$ and $a_{w_j}(o_i)$ are considered as the mapping probabilities in terms of objects and words, respectively.
We define the connection strength $e_{o_iw_j}$ as the product of these two probabilities.
Experimental results in Section \ref{sec:mapping} show that this cross-mapping approach makes associations between objects and words more accurate.

Finally, the cross-modal relational graph networks $G$ is constructed, and we can obtain the object-to-word mapping probability distributions $p(w_j|o_i)$ as shown in Equation \ref{eq:prob} using the object-word connection strengths.
%
\begin{equation}
\label{eq:prob}
\begin{gathered}
p(w_j|o_i) = {e_{o_iw_j}} /{\sum\limits_{w_r \in V^W} e_{o_iw_r}} 
\end{gathered}
\end{equation}
%
In the next section, we describe a method for learning an aligned cross-modal representation from the cross-modal graph $G$ where objects and words are aligned.

\section{Aligned cross-modal representations}

\begin{figure}[t]
\begin{center}
\centerline{\includegraphics[width=0.8\columnwidth]{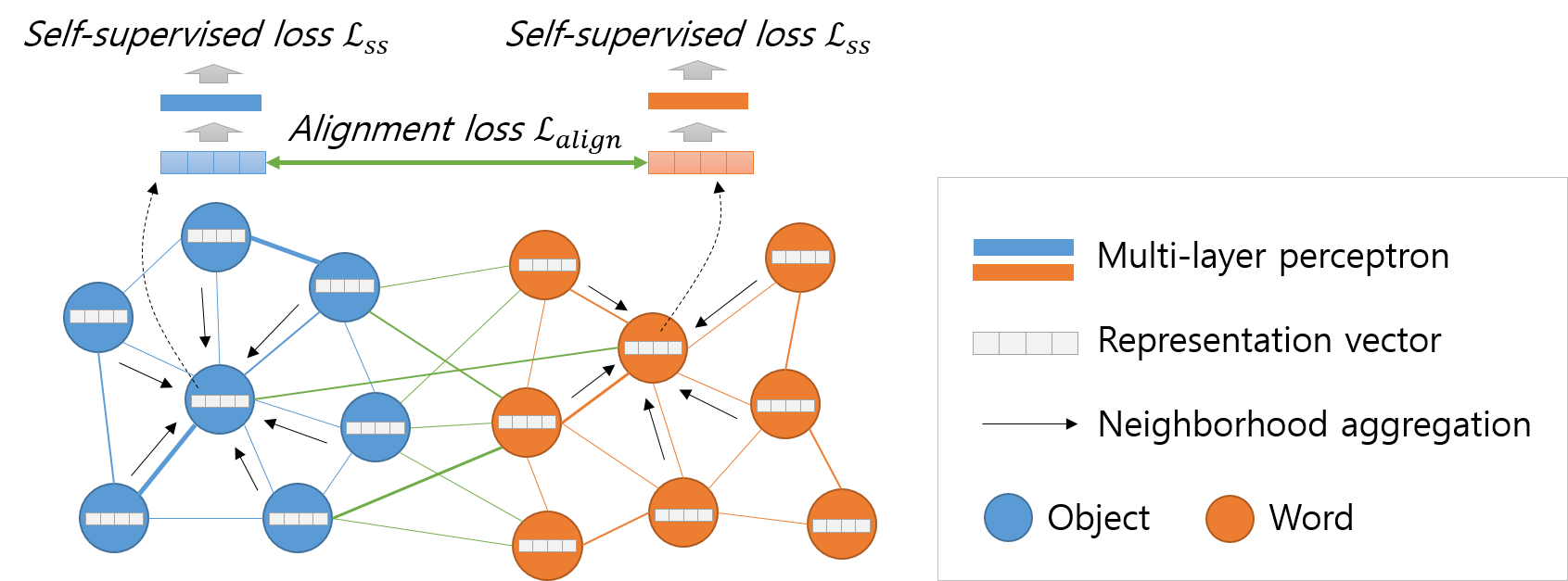}}
\caption{Overview of the aligned cross-modal representation learning method. First, we obtain a vector in which representation information of neighboring nodes is integrated using the neighborhood aggregation method.
We then use the aggregated vector to minimize alignment and self-supervised losses.
This process is performed for all nodes and edges of the graph networks.}
\label{fig:alignment_learning}
\end{center}
\vskip -0.2in
\end{figure}

We propose an aligned cross-modal representation learning method that learns the semantic representations of objects and words based on the constructed cross-modal relational graph networks.
It allows entities of different modalities with the same conceptual meaning to have the same representation vector.
Each object and word node has its own semantic representation vector $\vec{r}_{o_i}$ or $\vec{r}_{w_i}$, each initialized to a random vector.
The vectors are trained with the cross-entropy loss for self-identification of each node and the alignment loss for the correspondence between objects and words.

\subsection{Neighborhood aggregation}

Although object and word co-occurrence statistics were extracted from the local context, a neighborhood aggregation method can be used to reflect the global context.
The neighborhood aggregation function for cross-modal relational graph networks $G$ is defined as follows \cite{kim2021message}.
%
\begin{equation} \label{eq:aggregation}
\begin{gathered}
\vec{r}_{o_i}^{(l)} = \vec{r}_{o_i}^{(l-1)} + \delta \sum_{o_j \in \mathcal{N}_{o_i}} e_{o_io_j} \vec{r}_{o_j}^{(l-1)}, \ \forall o_i\in V^O \\
\vec{r}_{w_i}^{(l)} = \vec{r}_{w_i}^{(l-1)} + \delta \sum_{w_j \in \mathcal{N}_{w_i}} e_{w_iw_j} \vec{r}_{w_j}^{(l-1)}, \ \forall w_i\in V^W
\end{gathered}
\end{equation}
%
Here, $\delta$ is a constant between 0 and 1 to determine the propagation rate, and $\mathcal{N}_{o_i}$ and $\mathcal{N}_{w_i}$ is 1-hop neighbors of nodes $o_i$ and $w_i$, respectively.
By performing this process on the representation vectors $\vec{r}_{o_i}^{(l-1)}$ and $\vec{r}_{w_i}^{(l-1)}$ of all nodes, the updated vectors $\vec{r}_{o_i}^{(l)}$ and $\vec{r}_{w_i}^{(l)}$ for the next layer $l$ is obtained.
For layer number $l$, neighborhood aggregation is repeatedly performed from $l=1$ to $l=L$, and the base value $\vec{r}_{o_i}^{(0)}$ and $\vec{r}_{w_i}^{(0)}$ for layer $0$ are the node's own representation vectors $\vec{r}_{o_i}$ and $\vec{r}_{w_i}$.

As this method is repeatedly performed in multiple layers, information from more distant nodes can be aggregated.
Finally, we can train the representation vectors $\vec{r}_{o_i}$ and $\vec{r}_{w_i}$ of all nodes using the aggregated vectors $\vec{r}_{o_i}^{(L)}$ and $\vec{r}_{w_i}^{(L)}$ of the final layer $L$.

\subsection{Learning aligned cross-modal representations}

For each modality, we can train semantic representation vectors by inputting the aggregated vectors into a multi-layer perceptron (MLP) and performing self-supervised learning to classify the identity of each node.
More specifically, we use an MLP with a single hidden layer $h^O$ with no activation function for the object-relational graph networks $G^O$.
The dimension of the output layer of the MLP is set equal to the number of nodes in $G^O$, and the \textit{softmax function} is applied to the output to identify the input vector.
We also use an MLP with a hidden layer $h^W$ for word-relational graph networks $G^W$.
This method enables the semantic representation vectors to maintain their own identity while reflecting the information of neighboring nodes.

We add an alignment loss to the objective function $\mathcal{L}_{total}$ based on the connection strengths $e_{o_*w_*}$ of the cross-modal relational graph networks $G$.
The connection strength $e_{o_iw_j}$ between the object node $o_i$ and word node $w_j$ indicates the degree to which the meaning of the corresponding object and the word are similar.
We use the alignment loss term $\mathcal{L}_{align}$ to minimize the cosine distance between two semantic vectors by weighting the connection strength between object and word nodes.
Finally, the objective function $\mathcal{L}_{total}$ for the aligned representation learning is defined as Equation \ref{eq:align_loss} and \ref{eq:objective_function}.
%
\begin{equation} \label{eq:align_loss}
\mathcal{L}_{align}(\vec{r}_{o_i}^{(L)},\vec{r}_{w_j}^{(L)}) = e_{o_iw_j} \cdot  CosineDistance(\vec{r}_{o_i}^{(L)},\vec{r}_{w_j}^{(L)})
\end{equation}
\vskip -0.2in
\begin{equation} \label{eq:objective_function}
\begin{split}
\mathcal{L}_{total} &= \sum_{o_i \in V^O} \mathcal{L}_{ss}({z(\vec{r}_{o_i}^{(L)};h^O),o_i)} + \sum_{w_i \in V^W} \mathcal{L}_{ss}({z(\vec{r}_{w_i}^{(L)};h^W),w_i)}  \\
& + \sum_{o_i \in V^O} \sum_{w_j \in V^W} \lambda_{align} \mathcal{L}_{align}(\vec{r}_{o_i}^{(L)},\vec{r}_{w_j}^{(L)}) \\
\end{split}
\end{equation}
%
We write $z(\vec{r}_{o_i}^{(L)};h^O)$ and $z(\vec{r}_{w_i}^{(L)};h^W)$ as outputs of the MLP with hidden layers $h^O$ and $h^W$, with aggregate vectors $\vec{r}_{o_i}^{(L)}$ and $\vec{r}_{w_i}^{(L)}$ as inputs, respectively.
The self-supervised loss term $\mathcal{L}_{ss}$ is the standard softmax objective for self-identification.
The alignment loss term $\mathcal{L}_{align}(\vec{r}_{o_i}^{(L)},\vec{r}_{w_j}^{(L)})$ indicates cosine distance of two input vectors $\vec{r}_{o_i}^{(L)}$ and $\vec{r}_{w_j}^{(L)}$, where $\lambda_{align}$ is a hyperparameter.
The overall flow of the proposed method is shown in Figure \ref{fig:alignment_learning}.

\section{Experiments}
\label{sec:experiments}

We designed two tasks to investigate the effectiveness of the proposed model.
The object-to-word mapping task described in Section \ref{sec:object_mapping} evaluates the performance of the proposed XSL method.
Also, the zero-shot learning task in Section \ref{sec:zeroshot} validates the effectiveness of the aligned cross-modal representation learning method.
In all experiments, the size of the local context window $H$ was set to $10$, and the propagation rate $\delta$ and the number of layers $L$, which are parameters for neighborhood aggregation, were set to $0.3$ and $5$, respectively.
The value of $\lambda_{align}$ for the alignment loss term was set to 1.6, and the size of the embedding vector of each object and word was set to 100 dimensions.
We used the Adam optimizer \cite{kingma2014adam} to minimize the objective function $\mathcal{L}_{total}$ for training the representation vectors, and the learning rate and weight decay are set to 0.001 and 1e-5, respectively.
For experiments on hyperparameter settings, please refer to Appendix.

\subsection{Datasets}

We used the Open Image Dataset V6 \cite{OpenImages}, which consisted of approximately 1.9 million images containing 601 visual object categories and natural language descriptions.
We used verified image-level labels for all images, and on average, each image contained 3.8 object categories.
Each image description consists of an average of 38.3 tokenized words.
For efficient experiments, we removed words that appear less than five times in all descriptions, so there are a total of 6,209 words in the vocabulary.
The number of images without description is about 1.2 million. In this case, we only updated the object-relational graph networks using image-level pseudo labels.
Since we assumed a stream input environment, all data samples are input to the proposed model only once, one by one.

\subsection{Object-to-word mapping}
\label{sec:object_mapping}
We measured the mapping accuracy of objects and words for evaluation of the constructed cross-modal relational graph networks.
Accuracy assessments were performed on 369 objects whose names consisted of only one word and were included in the word vocabulary.
We did not limit the words that can map to objects to nouns, but used all words in the vocabulary: nouns, verbs, adjectives, adverbs, interrogatives, function words, etc.
Therefore, it is not an easy task to find the exact mapping of one of the 6,209 words for each object.
Mappings of the singular and plural with the same meaning (e.g., dog and dogs) were treated as true positives, and other cases such as synonyms were treated as incorrect mappings. 
We measured top-$K$ accuracy, which evaluates the accuracy of the $K$ most probable mappings predicted by the model.
As a baseline, we investigated a probabilistic method \cite{fazly2010probabilistic,kadar2015learning} `Prior' which defined as $e_{o_iw_j}=\nicefrac{p(o_i|w_j)}{{\sum_{w_u \in W} p(o_i|w_u)}}$ where $p(o_i|w_j)=\nicefrac{c_{o_iw_j}}{d_{w_j}}$.
We also performed ablation studies to verify the effectiveness of the cross-mapping approach described in Section \ref{sec:xsl}.
Methods `w/o $a_{o}$' and `w/o $a_{w}$' in Table \ref{tab:mapping_acc} use $e_{o_iw_j}=a_{w_j}(o_i)$ and $e_{o_iw_j}=a_{o_i}(w_j)$, respectively, instead of $e_{o_iw_j}=a_{o_i}(w_j) \cdot a_{w_j}(o_i)$ in Equation \ref{eq:cross_weight}.
Furthermore, to examine the applicability of the proposed method to datasets without categorical labels of objects, we performed experiments using an object detector instead of human verified pseudo-labels.
Here, we used YOLOv5l \cite{glenn_jocher_2020_4154370} pre-trained using the MS COCO dataset \cite{lin2014microsoft}.
It can detect 80 categories of objects, and the mapping accuracy was measured for 65 of these categories with single-word object names.

Finally, the continuous learning performance of the proposed model was verified by measuring the mapping accuracy according to the training samples and vocabulary increase.
Accuracy evaluation was performed every time 10k image samples were trained, and each time, the number of objects and word vocabularies learned by the model was counted.

\subsection{Zero-shot learning}
\label{sec:zeroshot}
The zero-shot learning task was designed to show that conceptual systems are topologically aligned even for never co-occurred object-word pairs.
We first set the cumulative number of co-occurrences $c_{o_iw_j}$ to 0 for the $Z$ correct object-word pairs ($o_i$-$w_j$) in the cross-modal relational graph networks.
We then trained the aligned cross-modal representations using the proposed method.

Zero-shot mapping accuracy was measured for 10 and 20 object-word pairs.
For a given $Z$ object-word pair, there are $Z!$ possible one-to-one mappings.
To find the best case among them, we selected the one with the largest sum of cosine similarities between the mapped representation vectors.
In this process, the Hungarian algorithm \cite{kuhn1955hungarian} was used to reduce the computation time.
The `w/o $\mathcal{L}_{align}$' method in Tabel \ref{tab:zero-shot5} is an ablation study to verify the effect of alignment loss in the proposed method, and was performed by removing the term $\mathcal{L}_{align}$ from the total loss $\mathcal{L}_{total}$ in Equation \ref{eq:objective_function}. 

As a baseline, we examined the mapping accuracy based on the alignment correlation (i.e., Spearman correlation) of the two conceptual systems \cite{horst2020conceptual}.
Here, Spearman correlations were obtained for up to 1M potential mappings randomly selected to find the best mapping with an acceptable computational cost.
We also measured random mapping accuracy for comparison.
For statistical significance, we repeated each experiment 30 times and $Z$ correct object-word pairs were randomly selected for each trial.
Finally, we qualitatively evaluated the proposed method by performing zero-shot learning on 100 correct object-word pairs and visualizing their similarity.

\begin{table}[t]
\caption{Object-to-word mapping accuracies from top-1 to top-5. (unit : \%)}
\label{tab:mapping_acc}
    \begin{center}
    \begin{small}
        \begin{adjustbox}{max width=\columnwidth}
        \begin{tabular}{c c c c c c c c c c c c c c}
        \toprule
        Pseudo label & Method & Top-1 & Top-2 & Top-3  & Top-4  & Top-5 & Pseudo label & Method & Top-1 & Top-2 & Top-3  & Top-4  & Top-5 \\
        \cmidrule(r){1-7} \cmidrule(r){8-14}
        & Prior & 37.13 & 50.68 & 57.18 & 60.98 & 64.23 & & Prior & 13.84 & 21.54 & 27.69 & 33.85 & 38.46 \\
        \cmidrule(r){2-7} \cmidrule(r){9-14}
        Human & w/o $a_{o}$ & 45.53 & 62.87 & 71.54 & 75.07 & 78.05 & Object & w/o $a_{o}$ & 38.46 & 52.31 & 60.00 & 67.69 & 78.46 \\
        verified & w/o $a_{w}$ & 70.19 & 79.13 & 82.11 & 84.28 & 85.91 & detector & w/o $a_{w}$ & 52.31 & 72.31 & 76.92 & 80.00 & 80.00 \\
        \cmidrule(r){2-7} \cmidrule(r){9-14}
        & Ours & \textbf{75.88} & \textbf{84.01} & \textbf{88.08} & \textbf{90.51} & \textbf{91.06} & & Ours & \textbf{70.77} & \textbf{86.15} & \textbf{92.31} & \textbf{95.38} & \textbf{96.92} \\
        \bottomrule
        \end{tabular}
        \end{adjustbox}
    \end{small}
    \end{center}
    \vskip -0.1in
\end{table}

\begin{table}[t]
\caption{Successful and unsuccessful cases of object-to-word mapping using human verified labels.}
\label{tab:mapping_top3}
    \begin{center}
    \begin{small}
    \begin{adjustbox}{max width=0.99\columnwidth}
        \begin{tabular}{c c c c c c c c c c}
        \toprule
         & Object & Word (top-3) & Prob. (\%) & Object  & Word (top-3) & Prob. (\%) & Object  & Word (top-3) & Prob. (\%) \\
        \cmidrule(r){1-4} \cmidrule(r){5-7} \cmidrule(r){8-10}
        & \multirow{3}{*}{porcupine}  &  porcupine  &  50.51 & \multirow{3}{*}{ambulance}  &  ambulance  &  85.68 & \multirow{3}{*}{eagle}  &  eagle  &  88.34 \\
        &  &  hedgehog  &  21.66 &   &  ambulances  &  13.45 &   &  flying  &  7.34 \\
        &  &  porcupines  &  19.07 &   &  stopped  &  0.53 & &  eagles  &  1.93 \\
        \cmidrule(r){2-4} \cmidrule(r){5-7} \cmidrule(r){8-10}
        Successful & \multirow{3}{*}{strawberry}  &  strawberries  &  69.31 & \multirow{3}{*}{watch}  &  watch  &  47.86 & \multirow{3}{*}{sink}  &  sink  &  33.48 \\
        cases &  &  strawberry  &  29.79 &   &  dial  &  19.31 & &  wash  &  16.69 \\
        &  &  raspberries  &  0.17 &   &  watches  &  11.71 & &  basin  &  14.99\\
        \cmidrule(r){2-4} \cmidrule(r){5-7} \cmidrule(r){8-10}
        & \multirow{3}{*}{train}  &  train  &  42.38 & \multirow{3}{*}{pen}  &  pen  &  48.03 & \multirow{3}{*}{cookie}  &  cookies  &  80.87 \\
        &  &  track  &  19.28 &   &  pens  &  24.63 & &  biscuits & 10.39 \\
        &  &  railway  &  17.75 &   &  ink  &  15.80 & &  cookie & 3.82 \\
        \midrule
        & \multirow{3}{*}{blender}  &  mixer  &  30.35 & \multirow{3}{*}{mouse}  &  rat  &  85.27 & \multirow{3}{*}{drink}  &  bottle & 17.67 \\
        \multirow{3}{*}{Unsuccessful} & &  juicer  &  26.23 & &  rats  &  7.06 & & wine  &  13.03 \\
        \multirow{3}{*}{cases} & &  grinder  &  22.85 & &  guinea  &  2.69 & & bottles  &  10.98 \\
        \cmidrule(r){2-4} \cmidrule(r){5-7} \cmidrule(r){8-10}
        & \multirow{3}{*}{seafood}  &  fish  &  13.97  & \multirow{3}{*}{animal}  &  bird  &  19.90 & \multirow{3}{*}{desk}  &  laptop  &  9.72 \\
        & &  prawns  &  10.65 & &  dog  &  6.56 & & computer  &  8.51\\
        & &  shells  &  8.30 & &  birds  &  4.61 & & monitor  &  7.71 \\
        \bottomrule
        \end{tabular}
        \end{adjustbox}
    \end{small}
    \end{center}
    \vskip -0.1in
\end{table}

\begin{figure}[h]
\vskip -0.02in
\begin{center}
\centerline{\includegraphics[width=0.8\columnwidth]{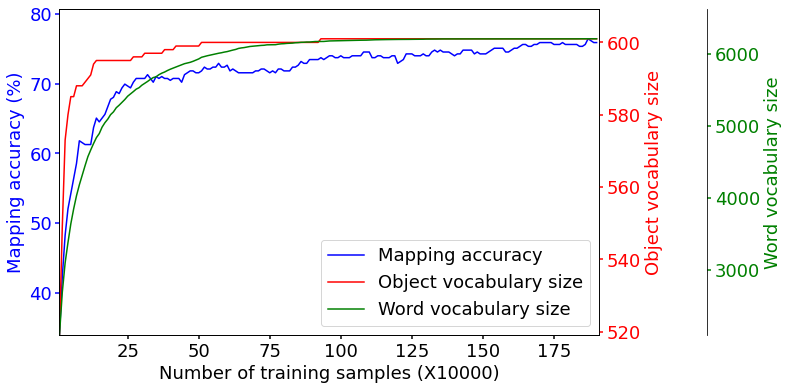}}
 \caption{Object-to-word mapping accuracy with increasing training samples. The red and green lines indicate the accumulated size of object and word vocabulary, respectively. As learning progresses, the size of vocabulary increases, and the mapping accuracy also tends to increase continuously.}
\label{fig:learning_progress}
\end{center}
\vskip -0.22in
\end{figure}

\section{Results and discussion}
\label{sec:results}

\subsection{Objects-to-words mapping}
\label{sec:mapping}
Table \ref{tab:mapping_acc} shows the top-1 to top-5 object-to-word mapping accuracies after training all data.
In all experiments, the accuracy of the proposed method (`Ours') was much higher than that of the baseline `Proir'.
Ours method also showed higher accuracy than methods `w/o $a_{o}$' and `w/o $a_{w}$', verifying that the proposed cross-mapping approach is effective.
Examples of successful top-3 object-to-word mappings are shown in Table \ref{tab:mapping_top3}.
In general, the mapping probability of the same word as the object name is high, and the related object name or verb is also included in the top-3.
For instance, pairs that are not accurately mapped, such as `train-track', `watch-dial', and `eagle-flying' are also composed of highly related ones.
Please refer to the Appendix for more examples.

Figure \ref{fig:learning_progress} shows the mapping accuracy as the number of learned samples increases.
As the size of the vocabulary increases rapidly at the beginning of training, the mapping accuracy also increases steeply.
Even after most of the vocabulary has been observed, the accuracy tends to steadily increase as the model is exposed to various situations.
It indicates that the proposed XSL method works properly without forgetting in the aspect of continuous learning.

\textbf{Limitations of evaluation methodology}\quad
We list unsuccessful cases of object-to-word mapping in Table \ref{tab:mapping_top3}.
Since we treated the mapping for synonyms as false positives, the most unsuccessful cases were mappings with words that have almost the same meaning as objects, such as `blender-mixer' and `mouse-rat'.
When the scope of the object name is inclusive, the hyponym was mapped in some cases, such as `seafood-fish' and `animal-bird'.
There are other cases such as `drink-bottle' and `desk-laptop', but most of them were mapped to words that are highly related to objects.

\subsection{Zero-shot learning}
\label{sec:zero_results}


Table \ref{tab:zero-shot5} shows the mapping accuracy of zero-shot learning task. 
For all tasks, the proposed method (`Ours') showed significantly higher mapping accuracy than `w/o $\mathcal{L}_{align}$' and other baselines. 
It confirms that objects and words with the same meaning have similar representation vectors in each conceptual system, even if object-word pairs are not directly learned.
Also, there was no significant difference in the accuracy of the 'Random' method and the `w/o $\mathcal{L}_{align}$' method, indicating that alignment loss played a decisive role for topologically aligned representation learning.

The perplexity of self-supervised loss $\mathcal{L}_{ss}$ was measured as low as less than 1.05.
Maintaining the identity of an entity's representation vector is an important aspect when using it for downstream tasks.
We used the neighborhood aggregation method to reflect the relationships of entities, and the alignment loss to make the representations of related objects and words similar.
These methods negatively affect the semantic representation of entities to maintain their identity.
Nevertheless, the learned representation retains its identity. 

\begin{figure}[t]
  \centering
  \begin{subfigure}[b]{0.52\textwidth}
    \includegraphics[width=\textwidth]{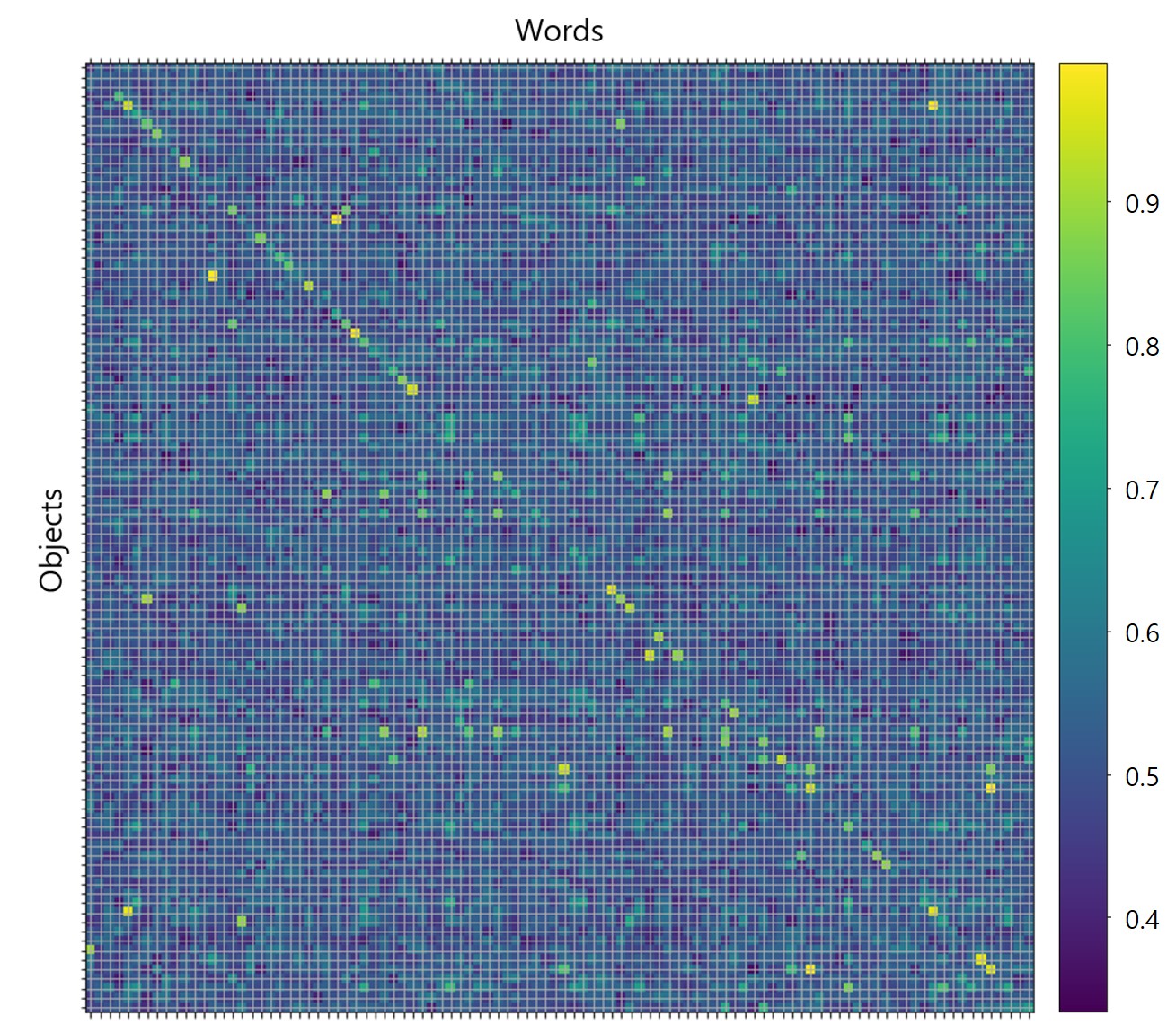}
  \end{subfigure}
  \hfill
  \begin{subfigure}[b]{0.445\textwidth}
    \includegraphics[width=\textwidth]{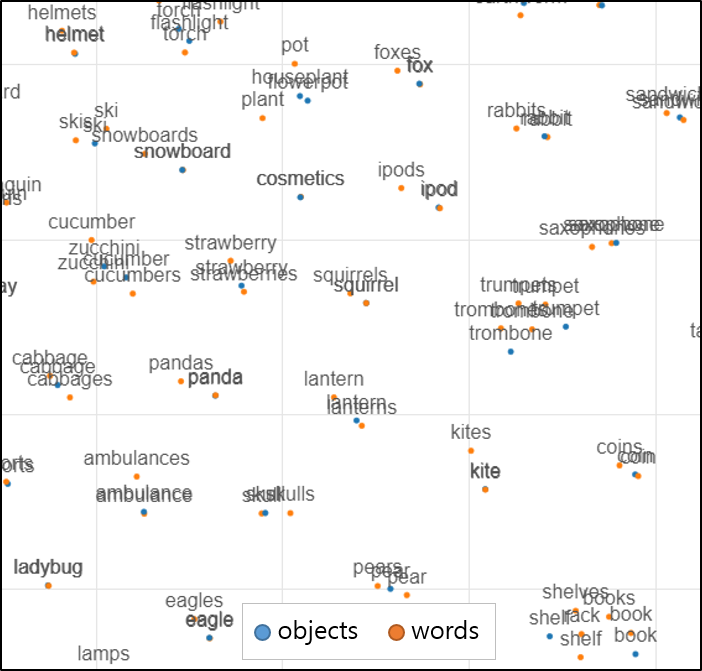}
  \end{subfigure}
  \caption{\textbf{Left:} Similarity matrix visualized by performing the zero-shot learning task on 100 object-word pairs. For correct object-word pairs, objects are arranged on the vertical axis and words are arranged on the horizontal axis in order. High similarity values are distributed along the diagonal axis, which is the correct object-word mapping combination. \textbf{Right:} The visualization of the representation vectors of the learned object and word using t-SNE \cite{maaten2008visualizing}. Objects and words with the same or similar meaning are distributed close to each other.}
  \label{fig:embedding_results}
  \vskip -0.2in
\end{figure}

Additionally, we performed a zero-shot learning task on 100 object-word pairs and visualized their similarity in Figure \ref{fig:embedding_results} (Left).
Although object-word co-occurrence information for 16.7\% of the total visual object vocabulary was removed, high similarity values were generally shown for objects and words with the same meaning.
This indicates that the constructed object-relational and word-relational graph networks are topologically aligned.

\begin{table}[t]
\caption{Mapping accuracy (mean $\pm$ std) of the zero-shot learning task. Perplexity (mean $\pm$ std) for self-identification was measured for all objects and words.}
\label{tab:zero-shot5}
    \begin{center}
    \begin{small}
    \begin{adjustbox}{max width=0.99\columnwidth}
        \begin{tabular}{c c c c c c c c}
        \toprule
        Number & \multirow{2}{*}{Method} & \multirow{2}{*}{Perplexity} & Mapping & Number & \multirow{2}{*}{Method} & \multirow{2}{*}{Perplexity} & Mapping \\
        of pairs & & & accuracy (\%) & of pairs & & & accuracy (\%) \\
        \cmidrule(r){1-4} \cmidrule(r){5-8}
        
        & Random & N/A & 13.00 $\pm$ 10.88 & & Random & N/A & 6.17 $\pm$ 8.27   \\
        \multirow{2.6}{*}{10} & Spearman correlation & 1.03 $\pm$ 0.02 & 10.67 $\pm$ 10.15 & \multirow{2.6}{*}{20} & Spearman correlation & 1.03 $\pm$ 0.02 & 7.17 $\pm$ 6.91 \\
        \cmidrule(r){2-4} \cmidrule(r){6-8}
        &  w/o $\mathcal{L}_{align}$ & 1.00 $\pm$ 0.01 & 13.33 $\pm$ 11.55 & & w/o $\mathcal{L}_{align}$ & 1.01 $\pm$ 0.03 & 5.67 $\pm$ 5.83  \\
        \cmidrule(r){2-4} \cmidrule(r){6-8}
        & Ours & 1.03 $\pm$ 0.02 &  \textbf{63.33} $\pm$ 18.07 & & Ours & 1.03 $\pm$ 0.02 &  \textbf{54.67} $\pm$ 15.70\\
        \bottomrule
        \end{tabular}
        \end{adjustbox}
    \end{small}
    \end{center}
    \vskip -0.1in
\end{table}

\section{Conclusions}
\label{sec:conclusions}
We propose a cross-modal alignment learning method that learns the association of objects and words online based on the co-occurrence statistics of vision-language data. 
The proposed model continuously constructs cross-modal relational graph networks using a cross-situational learning method.
By continuously adding and modifying nodes and edges of the graph networks, continuous learning is possible without specifying the number of entities to be learned in advance.

Additionally, we propose an aligned cross-modal representation learning method that learns semantic representations of objects and words from the constructed cross-modal relational graph networks through self-supervision.
By adding an alignment loss term to the objective function, objects and words with conceptually similar meanings have similar semantic representation vectors.

\textbf{Limitations and societal impact}\quad
Our work has limitations, such as not being able to map the names of objects composed of two or more words, and not learning the relationship between words such as synonym and hyponym.
Experiments in this study were performed using a vision-language dataset, but this does not mean that language is acquired only by audiovisual stimuli.
Our approach can be extended to various modalities in the future works.

{
\small
\bibliographystyle{plainnat}
\bibliography{references}

\begin{thebibliography}{39}
\providecommand{\natexlab}[1]{#1}
\providecommand{\url}[1]{\texttt{#1}}
\expandafter\ifx\csname urlstyle\endcsname\relax
  \providecommand{\doi}[1]{doi: #1}\else
  \providecommand{\doi}{doi: \begingroup \urlstyle{rm}\Url}\fi

\bibitem[Bergelson and Swingley(2012)]{bergelson20126}
Elika Bergelson and Daniel Swingley.
\newblock At 6--9 months, human infants know the meanings of many common nouns.
\newblock \emph{Proceedings of the National Academy of Sciences}, 109\penalty0
  (9):\penalty0 3253--3258, 2012.

\bibitem[Bird(2006)]{bird2006nltk}
Steven Bird.
\newblock Nltk: the natural language toolkit.
\newblock In \emph{Proceedings of the COLING/ACL 2006 Interactive Presentation
  Sessions}, pages 69--72, 2006.

\bibitem[Castrejon et~al.(2016)Castrejon, Aytar, Vondrick, Pirsiavash, and
  Torralba]{castrejon2016learning}
Lluis Castrejon, Yusuf Aytar, Carl Vondrick, Hamed Pirsiavash, and Antonio
  Torralba.
\newblock Learning aligned cross-modal representations from weakly aligned
  data.
\newblock In \emph{Proceedings of the IEEE conference on computer vision and
  pattern recognition}, pages 2940--2949, 2016.

\bibitem[Clerkin and Smith(2019)]{clerkin2019everyday}
Elizabeth~M Clerkin and Linda~B Smith.
\newblock The everyday statistics of objects and their names: How word learning
  gets its start.
\newblock In \emph{CogSci}, pages 240--246, 2019.

\bibitem[Deng et~al.(2018)Deng, Wu, Wu, Hu, Lyu, and Tan]{deng2018visual}
Chaorui Deng, Qi~Wu, Qingyao Wu, Fuyuan Hu, Fan Lyu, and Mingkui Tan.
\newblock Visual grounding via accumulated attention.
\newblock In \emph{Proceedings of the IEEE conference on computer vision and
  pattern recognition}, pages 7746--7755, 2018.

\bibitem[Fazly et~al.(2010)Fazly, Alishahi, and
  Stevenson]{fazly2010probabilistic}
Afsaneh Fazly, Afra Alishahi, and Suzanne Stevenson.
\newblock A probabilistic computational model of cross-situational word
  learning.
\newblock \emph{Cognitive Science}, 34\penalty0 (6):\penalty0 1017--1063, 2010.

\bibitem[Harris(1954)]{harris1954distributional}
Zellig~S Harris.
\newblock Distributional structure.
\newblock \emph{Word}, 10\penalty0 (2-3):\penalty0 146--162, 1954.

\bibitem[He and Arunachalam(2017)]{he2017word}
Angela~Xiaoxue He and Sudha Arunachalam.
\newblock Word learning mechanisms.
\newblock \emph{Wiley Interdisciplinary Reviews: Cognitive Science}, 8\penalty0
  (4):\penalty0 e1435, 2017.

\bibitem[Horst and Bird(2020)]{horst2020conceptual}
Jessica~S Horst and Chris~M Bird.
\newblock Conceptual systems align to aid concept learning.
\newblock \emph{Nature Machine Intelligence}, 2\penalty0 (2):\penalty0 92--93,
  2020.

\bibitem[Jia et~al.(2021)Jia, Yang, Xia, Chen, Parekh, Pham, Le, Sung, Li, and
  Duerig]{jia2021scaling}
Chao Jia, Yinfei Yang, Ye~Xia, Yi-Ting Chen, Zarana Parekh, Hieu Pham, Quoc~V
  Le, Yunhsuan Sung, Zhen Li, and Tom Duerig.
\newblock Scaling up visual and vision-language representation learning with
  noisy text supervision.
\newblock \emph{arXiv preprint arXiv:2102.05918}, 2021.

\bibitem[Jocher et~al.(2020)Jocher, Stoken, Borovec, NanoCode012,
  ChristopherSTAN, Changyu, Laughing, tkianai, Hogan, lorenzomammana, yxNONG,
  AlexWang1900, Diaconu, Marc, wanghaoyang0106, ml5ah, Doug, Ingham, Frederik,
  Guilhen, Hatovix, Poznanski, Fang, Yu, changyu98, Wang, Gupta, Akhtar,
  PetrDvoracek, and Rai]{glenn_jocher_2020_4154370}
Glenn Jocher, Alex Stoken, Jirka Borovec, NanoCode012, ChristopherSTAN, Liu
  Changyu, Laughing, tkianai, Adam Hogan, lorenzomammana, yxNONG, AlexWang1900,
  Laurentiu Diaconu, Marc, wanghaoyang0106, ml5ah, Doug, Francisco Ingham,
  Frederik, Guilhen, Hatovix, Jake Poznanski, Jiacong Fang, Lijun Yu,
  changyu98, Mingyu Wang, Naman Gupta, Osama Akhtar, PetrDvoracek, and Prashant
  Rai.
\newblock {ultralytics/yolov5: v3.1 - Bug Fixes and Performance Improvements},
  October 2020.
\newblock URL \url{https://doi.org/10.5281/zenodo.4154370}.

\bibitem[Jusczyk and Aslin(1995)]{jusczyk1995infants}
Peter~W Jusczyk and Richard~N Aslin.
\newblock Infants' detection of the sound patterns of words in fluent speech.
\newblock \emph{Cognitive psychology}, 29\penalty0 (1):\penalty0 1--23, 1995.

\bibitem[K{\'a}d{\'a}r et~al.(2015)K{\'a}d{\'a}r, Alishahi, and
  Chrupa{\l}a]{kadar2015learning}
{\'A}kos K{\'a}d{\'a}r, Afra Alishahi, and Grzegorz Chrupa{\l}a.
\newblock Learning word meanings from images of natural scenes.
\newblock \emph{Traitement Automatique des Langues}, 55\penalty0 (3), 2015.

\bibitem[Khorrami and R{\"a}s{\"a}nen(2021)]{khorrami2021can}
Khazar Khorrami and Okko R{\"a}s{\"a}nen.
\newblock Can phones, syllables, and words emerge as side-products of
  cross-situational audiovisual learning?--a computational investigation.
\newblock \emph{arXiv preprint arXiv:2109.14200}, 2021.

\bibitem[Kim et~al.(2021)Kim, Hwang, Lee, Kim, Choi, Lim, and
  Zhang]{kim2021message}
Taehyeong Kim, Injune Hwang, Hyundo Lee, Hyunseo Kim, Won-Seok Choi, Joseph~J
  Lim, and Byoung-Tak Zhang.
\newblock Message passing adaptive resonance theory for online active
  semi-supervised learning.
\newblock In \emph{International Conference on Machine Learning}, pages
  5519--5529. PMLR, 2021.

\bibitem[Kingma and Ba(2014)]{kingma2014adam}
Diederik~P Kingma and Jimmy Ba.
\newblock Adam: A method for stochastic optimization.
\newblock \emph{arXiv preprint arXiv:1412.6980}, 2014.

\bibitem[Kuhn(1955)]{kuhn1955hungarian}
Harold~W Kuhn.
\newblock The hungarian method for the assignment problem.
\newblock \emph{Naval research logistics quarterly}, 2\penalty0 (1-2):\penalty0
  83--97, 1955.

\bibitem[Kuznetsova et~al.(2020)Kuznetsova, Rom, Alldrin, Uijlings, Krasin,
  Pont-Tuset, Kamali, Popov, Malloci, Kolesnikov, Duerig, and
  Ferrari]{OpenImages}
Alina Kuznetsova, Hassan Rom, Neil Alldrin, Jasper Uijlings, Ivan Krasin, Jordi
  Pont-Tuset, Shahab Kamali, Stefan Popov, Matteo Malloci, Alexander
  Kolesnikov, Tom Duerig, and Vittorio Ferrari.
\newblock The open images dataset v4: Unified image classification, object
  detection, and visual relationship detection at scale.
\newblock \emph{IJCV}, 2020.

\bibitem[Li et~al.(2020)Li, Yin, Li, Zhang, Hu, Zhang, Wang, Hu, Dong, Wei,
  et~al.]{li2020oscar}
Xiujun Li, Xi~Yin, Chunyuan Li, Pengchuan Zhang, Xiaowei Hu, Lei Zhang, Lijuan
  Wang, Houdong Hu, Li~Dong, Furu Wei, et~al.
\newblock Oscar: Object-semantics aligned pre-training for vision-language
  tasks.
\newblock In \emph{European Conference on Computer Vision}, pages 121--137.
  Springer, 2020.

\bibitem[Lin et~al.(2014)Lin, Maire, Belongie, Hays, Perona, Ramanan,
  Doll{\'a}r, and Zitnick]{lin2014microsoft}
Tsung-Yi Lin, Michael Maire, Serge Belongie, James Hays, Pietro Perona, Deva
  Ramanan, Piotr Doll{\'a}r, and C~Lawrence Zitnick.
\newblock Microsoft coco: Common objects in context.
\newblock In \emph{European conference on computer vision}, pages 740--755.
  Springer, 2014.

\bibitem[Liu et~al.(2020)Liu, Wan, Zhu, and He]{liu2020learning}
Yongfei Liu, Bo~Wan, Xiaodan Zhu, and Xuming He.
\newblock Learning cross-modal context graph for visual grounding.
\newblock In \emph{Proceedings of the AAAI Conference on Artificial
  Intelligence}, volume~34, pages 11645--11652, 2020.

\bibitem[Maaten and Hinton(2008)]{maaten2008visualizing}
Laurens van~der Maaten and Geoffrey Hinton.
\newblock Visualizing data using t-sne.
\newblock \emph{Journal of machine learning research}, 9\penalty0
  (Nov):\penalty0 2579--2605, 2008.

\bibitem[Mandler(2007)]{mandler2007origins}
Jean~M Mandler.
\newblock On the origins of the conceptual system.
\newblock \emph{American Psychologist}, 62\penalty0 (8):\penalty0 741, 2007.

\bibitem[Nikolaus and Fourtassi(2021)]{nikolaus2021evaluating}
Mitja Nikolaus and Abdellah Fourtassi.
\newblock Evaluating the acquisition of semantic knowledge from
  cross-situational learning in artificial neural networks.
\newblock In \emph{Workshop on Cognitive Modeling and Computational
  Linguistics}, pages 200--210. Association for Computational Linguistics,
  2021.

\bibitem[Pennington et~al.(2014)Pennington, Socher, and
  Manning]{pennington2014glove}
Jeffrey Pennington, Richard Socher, and Christopher~D Manning.
\newblock Glove: Global vectors for word representation.
\newblock In \emph{Proceedings of the 2014 conference on empirical methods in
  natural language processing (EMNLP)}, pages 1532--1543, 2014.

\bibitem[Pont-Tuset et~al.(2020)Pont-Tuset, Uijlings, Changpinyo, Soricut, and
  Ferrari]{pont2020connecting}
Jordi Pont-Tuset, Jasper Uijlings, Soravit Changpinyo, Radu Soricut, and
  Vittorio Ferrari.
\newblock Connecting vision and language with localized narratives.
\newblock In \emph{European Conference on Computer Vision}, pages 647--664.
  Springer, 2020.

\bibitem[Quine(2013)]{quine2013word}
Willard Van~Orman Quine.
\newblock \emph{Word and object}.
\newblock MIT press, 2013.

\bibitem[Quinn et~al.(2001)Quinn, Eimas, and Tarr]{quinn2001perceptual}
Paul~C Quinn, Peter~D Eimas, and Michael~J Tarr.
\newblock Perceptual categorization of cat and dog silhouettes by 3-to
  4-month-old infants.
\newblock \emph{Journal of experimental child psychology}, 79\penalty0
  (1):\penalty0 78--94, 2001.

\bibitem[Radford et~al.(2021)Radford, Kim, Hallacy, Ramesh, Goh, Agarwal,
  Sastry, Askell, Mishkin, Clark, et~al.]{radford2021learning}
Alec Radford, Jong~Wook Kim, Chris Hallacy, Aditya Ramesh, Gabriel Goh,
  Sandhini Agarwal, Girish Sastry, Amanda Askell, Pamela Mishkin, Jack Clark,
  et~al.
\newblock Learning transferable visual models from natural language
  supervision.
\newblock \emph{arXiv preprint arXiv:2103.00020}, 2021.

\bibitem[R{\"a}s{\"a}nen and Rasilo(2015)]{rasanen2015joint}
Okko R{\"a}s{\"a}nen and Heikki Rasilo.
\newblock A joint model of word segmentation and meaning acquisition through
  cross-situational learning.
\newblock \emph{Psychological review}, 122\penalty0 (4):\penalty0 792, 2015.

\bibitem[Roads and Love(2020)]{roads2020learning}
Brett~D Roads and Bradley~C Love.
\newblock Learning as the unsupervised alignment of conceptual systems.
\newblock \emph{Nature Machine Intelligence}, 2\penalty0 (1):\penalty0 76--82,
  2020.

\bibitem[Roesler et~al.(2018)Roesler, Aly, Taniguchi, and
  Hayashi]{roesler2018probabilistic}
Oliver Roesler, Amir Aly, Tadahiro Taniguchi, and Yoshikatsu Hayashi.
\newblock A probabilistic framework for comparing syntactic and semantic
  grounding of synonyms through cross-situational learning.
\newblock In \emph{ICRA-2018 Workshop on" Representing a Complex World:
  Perception, Inference, and Learning for Joint Semantic, Geometric, and
  Physical Understanding"}, 2018.

\bibitem[Suanda et~al.(2014)Suanda, Mugwanya, and Namy]{suanda2014cross}
Sumarga~H Suanda, Nassali Mugwanya, and Laura~L Namy.
\newblock Cross-situational statistical word learning in young children.
\newblock \emph{Journal of experimental child psychology}, 126:\penalty0
  395--411, 2014.

\bibitem[Swingley(2009)]{swingley2009contributions}
Daniel Swingley.
\newblock Contributions of infant word learning to language development.
\newblock \emph{Philosophical Transactions of the Royal Society B: Biological
  Sciences}, 364\penalty0 (1536):\penalty0 3617--3632, 2009.

\bibitem[Taniguchi et~al.(2017)Taniguchi, Taniguchi, and
  Cangelosi]{taniguchi2017cross}
Akira Taniguchi, Tadahiro Taniguchi, and Angelo Cangelosi.
\newblock Cross-situational learning with bayesian generative models for
  multimodal category and word learning in robots.
\newblock \emph{Frontiers in neurorobotics}, 11:\penalty0 66, 2017.

\bibitem[Vlach and Johnson(2013)]{vlach2013memory}
Haley~A Vlach and Scott~P Johnson.
\newblock Memory constraints on infants' cross-situational statistical
  learning.
\newblock \emph{Cognition}, 127\penalty0 (3):\penalty0 375--382, 2013.

\bibitem[Xiao et~al.(2017)Xiao, Sigal, and Jae~Lee]{xiao2017weakly}
Fanyi Xiao, Leonid Sigal, and Yong Jae~Lee.
\newblock Weakly-supervised visual grounding of phrases with linguistic
  structures.
\newblock In \emph{Proceedings of the IEEE Conference on Computer Vision and
  Pattern Recognition}, pages 5945--5954, 2017.

\bibitem[Yu and Smith(2007)]{yu2007rapid}
Chen Yu and Linda~B Smith.
\newblock Rapid word learning under uncertainty via cross-situational
  statistics.
\newblock \emph{Psychological science}, 18\penalty0 (5):\penalty0 414--420,
  2007.

\bibitem[Yu et~al.(2018)Yu, Yu, Xiang, Zhao, Tian, and Tao]{yu2018rethinking}
Zhou Yu, Jun Yu, Chenchao Xiang, Zhou Zhao, Qi~Tian, and Dacheng Tao.
\newblock Rethinking diversified and discriminative proposal generation for
  visual grounding.
\newblock \emph{arXiv preprint arXiv:1805.03508}, 2018.

\end{thebibliography}
}

\newpage
\appendix
\section{Appendix}

\subsection{Dataset}

We used `Image labels' of the Open Image Dataset V6 \cite{OpenImages} with text descriptions from Localized Narratives \cite{pont2020connecting}.
Since the evaluation target of our experiments is the constructed cross-modal relational graph, we did not distinguish the split (i.e., train, validation, and test) of the provided dataset and used them all together.
Thus, the dataset consisted of a total of 1,908,119 images containing 601 visual object categories, of which text descriptions for 669,490 images were obtained from the Localized Narratives.
The number of images without descriptions is 1,238,629, and these samples were used only to update the object-relational graph.


We also performed experiments using a pre-trained object detector as pseudo labeler instead of human-verified labels to investigate the applicability of the proposed method.
We used YOLOv5l \cite{glenn_jocher_2020_4154370} as an object detector trained on the COCO dataset \cite{lin2014microsoft}.
The COCO dataset contains a total of 80 object categories for object detector training.
We validated the proposed method for 80 categories of pseudo-labels obtained using the object detector from image samples of Open Image V6.



The download link and license for each dataset are as follows.:
\begin{itemize}
\item Open Image Dataset V6: \url{https://storage.googleapis.com/openimages/web/download.html} (Apache License 2.0)
\item Localized Narratives: \url{https://google.github.io/localized-narratives/} (CC BY 4.0)
\item COCO: \url{https://cocodataset.org/} (CC BY 4.0)
\end{itemize}



\newpage

\subsection{Additional experimental results}

\subsubsection{Object-to-word mapping}
\textbf{Using human-verified labels} \quad
Table \ref{tab:human_successful} and \ref{tab:human_unsuccessful} show the object-to-word mapping results using human-verified pseudo-labels.

\begin{table}[ht]
\caption{Successful cases of object-to-word mapping using human-verified pseudo-labels. The mapping probability of the same word as the object name is generally high, and related object names or verbs are also included in the top-3.}
\label{tab:human_successful}
    \begin{center}
    \begin{small}
    \begin{adjustbox}{max width=0.99\columnwidth}
        \begin{tabular}{c c c c c c c c c}
        \toprule
        Object & Word (top-3) & Prob. (\%) & Object  & Word (top-3) & Prob. (\%) & Object  & Word (top-3) & Prob. (\%) \\
        \cmidrule(r){1-3} \cmidrule(r){4-6} \cmidrule(r){7-9}
        \multirow{3}{*}{dog}  &  dog  &  77.47 & \multirow{3}{*}{poster}  &  poster  &  71.62 & \multirow{3}{*}{table}  &  table  &  26.65 \\
          &  dogs  &  11.26 &   &  text  &  2.99 &   &  chairs  &  6.63 \\
          &  belt  &  3.91 &   &  images  &  2.34 &   &  tables  &  6.18 \\
        \cmidrule(r){1-3} \cmidrule(r){4-6} \cmidrule(r){7-9}
        \multirow{3}{*}{coconut}  &  coconuts  &  75.68 & \multirow{3}{*}{caterpillar}  &  caterpillar  &  90.03 & \multirow{3}{*}{sushi}  &  sushi  &  96.33 \\
          &  coconut  &  24.09 &   &  caterpillars  &  5.71 &   &  delicious  &  1.18 \\
          &  cocoa  &  0.13 &   &  worm  &  2.74 &   &  polaroid  &  0.35 \\
        \cmidrule(r){1-3} \cmidrule(r){4-6} \cmidrule(r){7-9}
        \multirow{3}{*}{coat}  &  coat  &  22.93 & \multirow{3}{*}{jeans}  &  jeans  &  43.78 & \multirow{3}{*}{pillow}  &  pillows  &  43.14 \\
          &  jacket  &  9.75 &   &  standing  &  3.71 &   &  bed  &  24.98 \\
          &  mannequin  &  8.77 &   &  walking  &  2.77 &   &  pillow  &  14.65 \\
        \cmidrule(r){1-3} \cmidrule(r){4-6} \cmidrule(r){7-9}
        \multirow{3}{*}{earrings}  &  earrings  &  87.31 & \multirow{3}{*}{vase}  &  vase  &  83.72 & \multirow{3}{*}{helicopter}  &  helicopter  &  86.20 \\
          &  earring  &  5.18 &   &  vases  &  6.00 &   &  helicopters  &  6.27 \\
          &  ear  &  4.75 &   &  pot  &  2.18 &   &  chopper  &  4.27 \\
        \cmidrule(r){1-3} \cmidrule(r){4-6} \cmidrule(r){7-9}
        \multirow{3}{*}{swan}  &  swan  &  63.25 & \multirow{3}{*}{broccoli}  &  broccoli  &  96.14 & \multirow{3}{*}{teapot}  &  teapot  &  65.17 \\
          &  swans  &  35.40 &   &  cauliflower  &  1.67 &   &  kettle  &  29.36 \\
          &  ducks  &  0.64 &   &  cauliflowers  &  1.14 &   &  teapots  &  2.15 \\
        \cmidrule(r){1-3} \cmidrule(r){4-6} \cmidrule(r){7-9}
        \multirow{3}{*}{seahorse}  &  seahorse  &  65.66 & \multirow{3}{*}{dishwasher}  &  dishwasher  &  66.92 & \multirow{3}{*}{dinosaur}  &  dinosaur  &  74.78 \\
          &  seahorses  &  32.21 &   &  mortar  &  22.38 &   &  skeleton  &  10.05 \\
          &  northern  &  1.46 &   &  electronics  &  4.83 &   &  dinosaurs  &  9.33 \\
        \cmidrule(r){1-3} \cmidrule(r){4-6} \cmidrule(r){7-9}
        \multirow{3}{*}{candy}  &  candies  &  48.91 & \multirow{3}{*}{cucumber}  &  cucumbers  &  58.42 & \multirow{3}{*}{porch}  &  house  &  27.35 \\
          &  candy  &  11.41 &   &  cucumber  &  38.08 &   &  porch  &  26.94 \\
          &  jellies  &  7.07 &   &  pickles  &  0.64 &   &  balcony  &  7.53 \\
        \cmidrule(r){1-3} \cmidrule(r){4-6} \cmidrule(r){7-9}
        \multirow{3}{*}{laptop}  &  laptop  &  67.09 & \multirow{3}{*}{door}  &  door  &  61.21 & \multirow{3}{*}{skirt}  &  skirt  &  80.75 \\
          &  laptops  &  25.11 &   &  doors  &  24.14 &   &  skirts  &  4.85 \\
          &  working  &  1.80 &   &  entrance  &  2.62 &   &  chalkboard  &  3.54 \\
        \cmidrule(r){1-3} \cmidrule(r){4-6} \cmidrule(r){7-9}
        \multirow{3}{*}{chair}  &  chairs  &  25.66 & \multirow{3}{*}{rocket}  &  rocket  &  67.82 & \multirow{3}{*}{football}  &  football  &  70.96 \\
          &  chair  &  23.19 &   &  rockets  &  12.96 &   &  ball  &  12.34 \\
          &  tables  &  5.54 &   &  launch  &  4.12 &   &  kicking  &  4.13 \\
        \cmidrule(r){1-3} \cmidrule(r){4-6} \cmidrule(r){7-9}
        \multirow{3}{*}{dragonfly}  &  dragonfly  &  89.73 & \multirow{3}{*}{clock}  &  clock  &  80.47 & \multirow{3}{*}{girl}  &  girl  &  29.44 \\
          &  dragonflies  &  4.90 &   &  clocks  &  9.28 &   &  girls  &  5.88 \\
          &  fly  &  3.21 &   &  seconds  &  1.37 &   &  woman  &  4.86 \\
        \cmidrule(r){1-3} \cmidrule(r){4-6} \cmidrule(r){7-9}
        \multirow{3}{*}{bear}  &  bear  &  82.39 & \multirow{3}{*}{couch}  &  sofa  &  51.19 & \multirow{3}{*}{cheetah}  &  cheetah  &  52.34 \\
          &  bears  &  8.63 &   &  couch  &  19.13 &   &  leopard  &  25.52 \\
          &  panda  &  5.16 &   &  sofas  &  10.16 &   &  leopards  &  12.83 \\
        \cmidrule(r){1-3} \cmidrule(r){4-6} \cmidrule(r){7-9}
        \multirow{3}{*}{toy}  &  toy  &  37.69 & \multirow{3}{*}{snowmobile}  &  snowmobile  &  58.31 & \multirow{3}{*}{shower}  &  shower  &  76.30 \\
          &  toys  &  37.27 &   &  snowmobiles  &  40.29 &   &  showers  &  8.36 \\
          &  lego  &  8.49 &   &  snow  &  0.36 &   &  washroom  &  3.26 \\
        \cmidrule(r){1-3} \cmidrule(r){4-6} \cmidrule(r){7-9}
        \multirow{3}{*}{rifle}  &  gun  &  51.58 & \multirow{3}{*}{pig}  &  pig  &  57.43 & \multirow{3}{*}{plant}  &  plants  &  20.56 \\
          &  rifle  &  17.08 &   &  pigs  &  34.42 &   &  grass  &  8.29 \\
          &  guns  &  9.94 &   &  piglets  &  5.45 &   &  trees  &  6.78 \\
        \cmidrule(r){1-3} \cmidrule(r){4-6} \cmidrule(r){7-9}
        \multirow{3}{*}{wheelchair}  &  wheelchair  &  55.54 & \multirow{3}{*}{tank}  &  tank  &  19.63 & \multirow{3}{*}{accordion}  &  accordion  &  64.69 \\
          &  wheelchairs  &  41.58 &   &  tanker  &  17.39 &   &  harmonium  &  24.71 \\
          &  wheel  &  2.16 &   &  war  &  12.01 &   &  accordions  &  6.95 \\
        \cmidrule(r){1-3} \cmidrule(r){4-6} \cmidrule(r){7-9}
        \multirow{3}{*}{backpack}  &  backpack  &  66.61 & \multirow{3}{*}{suit}  &  suit  &  24.59 & \multirow{3}{*}{tent}  &  tent  &  46.10 \\
          &  backpacks  &  7.93 &   &  blazer  &  12.98 &   &  tents  &  28.55 \\
          &  bag  &  5.37 &   &  suits  &  10.87 &   &  camping  &  19.20 \\
        \cmidrule(r){1-3} \cmidrule(r){4-6} \cmidrule(r){7-9}
        \multirow{3}{*}{tree}  &  trees  &  25.85 & \multirow{3}{*}{food}  &  food  &  28.22 & \multirow{3}{*}{cheese}  &  cheese  &  93.57 \\
          &  sky  &  9.35 &   &  plate  &  12.08 &   &  pizza  &  1.12 \\
          &  the  &  3.98 &   &  bowl  &  8.90 &   &  stuffings  &  0.99 \\
        \cmidrule(r){1-3} \cmidrule(r){4-6} \cmidrule(r){7-9}
        \multirow{3}{*}{lipstick}  &  lipstick  &  67.53 & \multirow{3}{*}{house}  &  house  &  28.64 & \multirow{3}{*}{suitcase}  &  suitcase  &  48.13 \\
          &  lipsticks  &  19.76 &   &  houses  &  10.60 &   &  luggage  &  26.43 \\
          &  lips  &  5.15 &   &  windows  &  8.97 &   &  suitcases  &  14.47 \\
        \bottomrule
        \end{tabular}
        \end{adjustbox}
    \end{small}
    \end{center}
\end{table}

\newpage
\begin{table}[h]
\caption{Unsuccessful cases of object-to-word mapping using human-verified pseudo-labels. Although the mapping is not exact, words with similar or related meanings are mostly ranked.}
\label{tab:human_unsuccessful}
    \begin{center}
    \begin{small}
    \begin{adjustbox}{max width=0.99\columnwidth}
        \begin{tabular}{c c c c c c c c c}
        \toprule
        Object & Word (top-3) & Prob. (\%) & Object  & Word (top-3) & Prob. (\%) & Object  & Word (top-3) & Prob. (\%) \\
        \cmidrule(r){1-3} \cmidrule(r){4-6} \cmidrule(r){7-9}
        \multirow{3}{*}{fax}  &  xerox  &  44.18 & \multirow{3}{*}{platter}  &  plates  &  25.84 & \multirow{3}{*}{mouse}  &  rat  &  85.27 \\
          &  photocopies  &  29.17 &   &  forks  &  17.79 &   &  rats  &  7.06 \\
          &  answering  &  23.67 &   &  plate  &  9.72 &   &  guinea  &  2.69 \\
        \cmidrule(r){1-3} \cmidrule(r){4-6} \cmidrule(r){7-9}
        \multirow{3}{*}{person}  &  the  &  5.87 & \multirow{3}{*}{antelope}  &  deer  &  41.97 & \multirow{3}{*}{seafood}  &  fish  &  13.97 \\
          &  a  &  5.57 &   &  impala  &  18.28 &   &  prawns  &  10.65 \\
          &  and  &  4.87 &   &  deers  &  15.81 &   &  shells  &  8.30 \\
        \cmidrule(r){1-3} \cmidrule(r){4-6} \cmidrule(r){7-9}
        \multirow{3}{*}{dessert}  &  cake  &  22.26 & \multirow{3}{*}{briefcase}  &  suitcase  &  32.43 & \multirow{3}{*}{trousers}  &  jeans  &  16.67 \\
          &  food  &  10.15 &   &  handler  &  14.15 &   &  pant  &  11.50 \\
          &  plate  &  8.54 &   &  leather  &  13.27 &   &  standing  &  4.05 \\
        \cmidrule(r){1-3} \cmidrule(r){4-6} \cmidrule(r){7-9}
        \multirow{3}{*}{canoe}  &  paddles  &  25.04 & \multirow{3}{*}{glasses}  &  spectacles  &  59.37 & \multirow{3}{*}{drink}  &  bottle  &  17.67 \\
          &  boat  &  18.03 &   &  specs  &  8.49 &   &  wine  &  13.03 \\
          &  rowing  &  15.25 &   &  spectacle  &  8.26 &   &  bottles  &  10.98 \\
        \cmidrule(r){1-3} \cmidrule(r){4-6} \cmidrule(r){7-9}
        \multirow{3}{*}{snack}  &  food  &  16.58 & \multirow{3}{*}{cattle}  &  cows  &  43.86 & \multirow{3}{*}{kitchenware}  &  thermos  &  84.39 \\
          &  packet  &  8.10 &   &  cow  &  37.39 &   &  headscarf  &  13.25 \\
          &  packets  &  7.39 &   &  buffalo  &  4.24 &   &  juices  &  0.67 \\
        \cmidrule(r){1-3} \cmidrule(r){4-6} \cmidrule(r){7-9}
        \multirow{3}{*}{heater}  &  burning  &  42.45 & \multirow{3}{*}{wheel}  &  car  &  18.68 & \multirow{3}{*}{container}  &  purifier  &  38.88 \\
          &  flame  &  24.17 &   &  road  &  9.35 &   &  fluid  &  28.44 \\
          &  firewood  &  13.77 &   &  cars  &  7.05 &   &  headscarf  &  18.28 \\
        \cmidrule(r){1-3} \cmidrule(r){4-6} \cmidrule(r){7-9}
        \multirow{3}{*}{countertop}  &  kitchen  &  25.88 & \multirow{3}{*}{clothing}  &  a  &  5.66 & \multirow{3}{*}{blender}  &  mixer  &  30.35 \\
          &  cupboards  &  12.79 &   &  the  &  5.43 &   &  juicer  &  26.23 \\
          &  sink  &  8.87 &   &  and  &  4.71 &   &  grinder  &  22.85 \\
        \cmidrule(r){1-3} \cmidrule(r){4-6} \cmidrule(r){7-9}
        \multirow{3}{*}{houseplant}  &  pots  &  28.59 & \multirow{3}{*}{desk}  &  laptop  &  9.72 & \multirow{3}{*}{flowerpot}  &  pots  &  38.47 \\
          &  pot  &  26.71 &   &  computer  &  8.51 &   &  pot  &  37.99 \\
          &  potted  &  10.99 &   &  monitor  &  7.71 &   &  potted  &  8.80 \\
        \cmidrule(r){1-3} \cmidrule(r){4-6} \cmidrule(r){7-9}
        \multirow{3}{*}{footwear}  &  shoes  &  8.24 & \multirow{3}{*}{cocktail}  &  juice  &  18.17 & \multirow{3}{*}{handgun}  &  pistol  &  39.71 \\
          &  the  &  5.05 &   &  straw  &  16.25 &   &  gun  &  24.58 \\
          &  people  &  3.55 &   &  cubes  &  14.51 &   &  revolver  &  15.78 \\
        \cmidrule(r){1-3} \cmidrule(r){4-6} \cmidrule(r){7-9}
        \multirow{3}{*}{shelf}  &  racks  &  25.06 & \multirow{3}{*}{chicken}  &  hen  &  49.20 & \multirow{3}{*}{beehive}  &  bees  &  46.75 \\
          &  shelves  &  23.67 &   &  hens  &  24.64 &   &  honeycomb  &  19.47 \\
          &  books  &  18.60 &   &  cock  &  11.30 &   &  honey  &  10.17 \\
        \cmidrule(r){1-3} \cmidrule(r){4-6} \cmidrule(r){7-9}
        \multirow{3}{*}{ladle}  &  use  &  45.19 & \multirow{3}{*}{tire}  &  car  &  14.47 & \multirow{3}{*}{scale}  &  weighing  &  79.95 \\
          &  scrub  &  32.37 &   &  vehicle  &  8.78 &   &  weight  &  9.19 \\
          &  droplet  &  14.31 &   &  tyre  &  8.49 &   &  balance  &  3.45 \\
        \cmidrule(r){1-3} \cmidrule(r){4-6} \cmidrule(r){7-9}
        \multirow{3}{*}{cream}  &  cosmetic  &  41.00 & \multirow{3}{*}{watercraft}  &  boats  &  25.97 & \multirow{3}{*}{shellfish}  &  shells  &  65.53 \\
          &  ointment  &  26.60 &   &  boat  &  22.40 &   &  seashells  &  11.22 \\
          &  tubes  &  13.43 &   &  ship  &  20.65 &   &  shell  &  5.61 \\
        \cmidrule(r){1-3} \cmidrule(r){4-6} \cmidrule(r){7-9}
        \multirow{3}{*}{bomb}  &  greenhouse  &  0.00 & \multirow{3}{*}{chisel}  &  blades  &  48.22 & \multirow{3}{*}{animal}  &  bird  &  19.90 \\
          &  signing  &  0.00 &   &  tool  &  47.11 &   &  dog  &  6.56 \\
          &  generator  &  0.00 &   &  tools  &  3.28 &   &  birds  &  4.61 \\
        \cmidrule(r){1-3} \cmidrule(r){4-6} \cmidrule(r){7-9}
        \multirow{3}{*}{furniture}  &  table  &  10.17 & \multirow{3}{*}{cosmetics}  &  nail  &  38.47 & \multirow{3}{*}{closet}  &  hangers  &  70.69 \\
          &  chair  &  8.06 &   &  polish  &  35.55 &   &  textile  &  6.99 \\
          &  chairs  &  7.88 &   &  nails  &  10.10 &   &  wardrobe  &  3.73 \\
        \cmidrule(r){1-3} \cmidrule(r){4-6} \cmidrule(r){7-9}
        \multirow{3}{*}{wok}  &  pan  &  29.46 & \multirow{3}{*}{mammal}  &  the  &  6.79 \\
          &  cooking  &  20.87 &   &  see  &  5.60 \\
          &  pans  &  14.49 &   &  in  &  5.50 \\
        \bottomrule
        \end{tabular}
        \end{adjustbox}
    \end{small}
    \end{center}
\end{table}

\newpage
\textbf{Using a pre-trained object detector} \quad
Table \ref{tab:detector_successful} and \ref{tab:detector_unsuccessful} show the object-to-word mapping results using the YOLOv5l \cite{glenn_jocher_2020_4154370} object detector as a pseudo labeler. 

\begin{table}[ht]
\caption{Successful cases of object-to-word mapping using the YOLOv5l object detector as a pseudo labeler. The mapping probability of the same word as the object name is generally high, and related object names or verbs are also included in the top-3.}
\label{tab:detector_successful}
    \begin{center}
    \begin{small}
    \begin{adjustbox}{max width=0.99\columnwidth}
        \begin{tabular}{c c c c c c c c c}
        \toprule
        Object & Word (top-3) & Prob. (\%) & Object  & Word (top-3) & Prob. (\%) & Object  & Word (top-3) & Prob. (\%) \\
        \cmidrule(r){1-3} \cmidrule(r){4-6} \cmidrule(r){7-9}
        \multirow{3}{*}{bicycle}  &  bicycle  &  46.38 & \multirow{3}{*}{car}  &  car  & 16.33 & \multirow{3}{*}{motorcycle}  &  bike  &  42.41 \\
         &  bicycles  &  23.28 &   &  cars  &  11.65 &   &  bikes  &  15.45 \\
         &  riding  &  10.06 &   &  vehicles  &  11.59 & &  motorcycle  &  9.35 \\
        \cmidrule(r){1-3} \cmidrule(r){4-6} \cmidrule(r){7-9}
         \multirow{3}{*}{airplane}  &  aircraft  &  23.80 & \multirow{3}{*}{bus}  &  bus  &  65.65 & \multirow{3}{*}{train}  &  train  &  38.92 \\
         &  airplane  & 19.25 &   &  buses  &  19.77 & &  track  &  19.61 \\
         &  runway  &  7.90 &   &  decker  &  3.51 & &  railway  &  18.48\\
        \cmidrule(r){1-3} \cmidrule(r){4-6} \cmidrule(r){7-9}
        \multirow{3}{*}{truck}  &  vehicles  &  16.23 & \multirow{3}{*}{boat}  &  boat  &  23.51 & \multirow{3}{*}{bench}  &  bench  &  49.00 \\
         &  vehicle  &  15.83 &   &  boats  &  23.43 & &  benches & 32.88 \\
         &  truck  &  10.00 &   &  water  &  13.39 & &  the & 0.97 \\
        \cmidrule(r){1-3} \cmidrule(r){4-6} \cmidrule(r){7-9}
        \multirow{3}{*}{bird}  &  bird  &  40.30 & \multirow{3}{*}{cat}  &  cat  &  84.58 & \multirow{3}{*}{dog}  &  dog  &  72.70 \\
         &  birds  &  14.58 &   &  cats  &  9.53 & &  dogs & 11.88 \\
         &  branch  &  3.72 &   &  rat  &  1.17 & &  belt & 4.69 \\
        \cmidrule(r){1-3} \cmidrule(r){4-6} \cmidrule(r){7-9}
        \multirow{3}{*}{horse}  &  horse  &  62.99 & \multirow{3}{*}{sheep}  &  sheep  &  52.74 & \multirow{3}{*}{cow}  &  cows  &  23.61 \\
         &  horses  &  25.44 &   &  goats  &  9.49 & &  cow & 19.08 \\
         &  cart  &  3.96 &   &  animals  &  9.41 & &  bull & 13.75 \\
        \cmidrule(r){1-3} \cmidrule(r){4-6} \cmidrule(r){7-9}
        \multirow{3}{*}{elephant}  &  elephant  &  58.63 & \multirow{3}{*}{zebra}  &  zebras  &  60.72 & \multirow{3}{*}{giraffe}  &  giraffe  &  55.38 \\
         &  elephants  &  31.42 &   &  zebra  &  23.83 & &  giraffes  &  30.60 \\
         &  rhinoceros  &  3.21 &   &  tiger  &  9.95 & &  cheetah  &  6.53 \\
        \cmidrule(r){1-3} \cmidrule(r){4-6} \cmidrule(r){7-9}
        \multirow{3}{*}{backpack}  &  bags  &  13.55 & \multirow{3}{*}{umbrella}  &  umbrella  &  37.10 & \multirow{3}{*}{tie}  &  tie  &  13.90 \\
         &  backpack  &  12.62 &   &  umbrellas  &  35.24 & &  suit  &  11.22 \\
         &  bag  &  8.57 &   &  tents  &  9.29 & &  blazer  &  5.76 \\
        \cmidrule(r){1-3} \cmidrule(r){4-6} \cmidrule(r){7-9}
        \multirow{3}{*}{suitcase}  &  luggage  &  28.17 & \multirow{3}{*}{frisbee}  &  frisbee  &  53.29 & \multirow{3}{*}{skis}  &  skiing  &  33.43 \\
         &  suitcase  &  21.90 &   &  disc  &  9.33 & &  ski  &  17.41 \\
         &  suitcases  &  10.58 &   &  weights  &  2.74 & &  skis  &  15.44 \\
        \cmidrule(r){1-3} \cmidrule(r){4-6} \cmidrule(r){7-9}
        \multirow{3}{*}{snowboard}  &  snowboard  &  40.81 & \multirow{3}{*}{kite}  &  parachute  &  28.16 & \multirow{3}{*}{skateboard}  &  skating  &  41.10 \\
         &  snow  &  25.43 &   &  kite  &  18.86 & &  skateboard  &  37.74 \\
         &  snowboarding  &  11.59 &   &  parachutes  &  14.90 & &  skate  &  8.15 \\
        \cmidrule(r){1-3} \cmidrule(r){4-6} \cmidrule(r){7-9}
        \multirow{3}{*}{surfboard}  &  surfing  &  45.23 & \multirow{3}{*}{bottle}  &  bottle  &  26.99 & \multirow{3}{*}{cup}  &  cup  &  8.57 \\
         &  surfboard  &  27.92 &   &  bottles  &  25.66 & &  table  &  8.24 \\
         &  surfboards  &  5.36 &   &  table  &  3.08 & &  glasses  &  6.80 \\
        \cmidrule(r){1-3} \cmidrule(r){4-6} \cmidrule(r){7-9}
        \multirow{3}{*}{fork}  &  fork  &  56.69 & \multirow{3}{*}{knife}  &  knife  &  50.50 & \multirow{3}{*}{spoon}  &  spoon  &  49.17 \\
         &  forks  &  14.31 &   &  knives  &  13.11 & &  spoons  &  9.46 \\
         &  spoons  &  5.28 &   &  spoons  &  4.44 & &  bowl  &  4.80 \\
        \cmidrule(r){1-3} \cmidrule(r){4-6} \cmidrule(r){7-9}
        \multirow{3}{*}{bowl}  &  bowl  &  33.45 & \multirow{3}{*}{banana}  &  bananas  &  57.04 & \multirow{3}{*}{apple}  &  fruits  &  35.01 \\
         &  bowls  &  14.78 &   &  banana  &  21.11 & &  apples  &  20.46 \\
         &  food  &  8.83 &   &  artichokes  &  1.76 & &  apple  &  10.09 \\
        \cmidrule(r){1-3} \cmidrule(r){4-6} \cmidrule(r){7-9}
        \multirow{3}{*}{orange}  &  oranges  &  22.56 & \multirow{3}{*}{broccoli}  &  broccoli  &  56.21 & \multirow{3}{*}{carrot}  &  carrot  &  29.52 \\
         &  fruits  &  20.21 &   &  cauliflower  &  4.76 & &  carrot  &  15.21 \\
         &  lemon  &  13.01 &   &  vegetables  &  3.48 & &  vegetables  &  7.91 \\
        \cmidrule(r){1-3} \cmidrule(r){4-6} \cmidrule(r){7-9}
        \multirow{3}{*}{pizza}  &  pizza  &  73.74 & \multirow{3}{*}{donut}  &  doughnuts  &  18.32 & \multirow{3}{*}{cake}  &  cake  &  58.80 \\
         &  pizzas  &  5.81 &   &  cookies  &  17.59  & &  cupcakes  &  12.86 \\
         &  food  &  5.23 &   &  donuts  &  12.93 & &  cakes  &  5.69 \\
        \cmidrule(r){1-3} \cmidrule(r){4-6} \cmidrule(r){7-9}
        \multirow{3}{*}{chair}  &  chairs  &  12.71 & \multirow{3}{*}{couch}  &  sofa  &  48.10  & \multirow{3}{*}{bed}  &  bed  &  60.00 \\
         &  chair  &  7.36 &   & couch  &  21.80  & &  pillows  &  10.46 \\
         &  sitting  &  5.75 &   &  sofas  &  8.89 & &  beds  &  6.37 \\
        \cmidrule(r){1-3} \cmidrule(r){4-6} \cmidrule(r){7-9}
        \multirow{3}{*}{toilet}  &  toilet  &  54.72 & \multirow{3}{*}{laptop}  &  laptop  &  50.09  & \multirow{3}{*}{mouse}  &  mouse  &  62.09 \\
         &  flush  &  9.93 &   & laptops  &  17.73  & &  keyboard  &  5.99 \\
         &  commode  &  7.87 &   &  working  &  1.94 & &  mouses  &  5.12 \\
        \cmidrule(r){1-3} \cmidrule(r){4-6} \cmidrule(r){7-9}
        \multirow{3}{*}{remote}  &  remote  &  52.08 & \multirow{3}{*}{keyboard}  &  keyboard  &  49.20  & \multirow{3}{*}{microwave}  &  oven  &  50.45 \\
         &  remotes  &  12.17 &   & mouse  &  8.23  & &  microwave  &  29.45 \\
         &  joystick  &  5.79 &   &  keyboards  &  7.00 & &  kitchen  &  3.98 \\
        \bottomrule
        \end{tabular}
        \end{adjustbox}
    \end{small}
    \end{center}
\end{table}

\newpage
\begin{table}[ht]
\caption{Unsuccessful cases of object-to-word mapping using the YOLOv5l object detector as a pseudo labeler. Although the mapping is not exact, words with similar or related meanings are mostly ranked.}
\label{tab:detector_unsuccessful}
    \begin{center}
    \begin{small}
    \begin{adjustbox}{max width=0.99\columnwidth}
        \begin{tabular}{c c c c c c c c c}
        \toprule
         Object & Word (top-3) & Prob. (\%) & Object  & Word (top-3) & Prob. (\%) & Object  & Word (top-3) & Prob. (\%) \\
        \cmidrule(r){1-3} \cmidrule(r){4-6} \cmidrule(r){7-9}
        \multirow{3}{*}{person}  &  the  &  6.84 & \multirow{3}{*}{bear}  &  animal  &  16.28 & \multirow{3}{*}{handbag}  &  bag  &  8.34 \\
        &  a  &  5.18 & &  monkey  &  11.61 & & people  &  5.37 \\
         &  in  &  4.85 & &  polar  &  10.66 & & bags  &  4.10 \\
        \cmidrule(r){1-3} \cmidrule(r){4-6} \cmidrule(r){7-9}
        \multirow{3}{*}{sandwich}  &  burger  &  27.05  & \multirow{3}{*}{tv}  &  screen  &  18.55 &  &  &  \\
        &  food  &  8.94 & &  television  &  16.89 & & & \\
        &  bread  &  8.69 & &  monitor  &  6.83 & &   & \\
        \bottomrule
        \end{tabular}
        \end{adjustbox}
    \end{small}
    \end{center}
\end{table}

\newpage
\subsubsection{Zero-shot learning}

\textbf{Parameter search} \quad
We searched the hyperparameters for training the aligned cross-modal representations in the space $\delta \in [0,0.5]$, $L \in [0,6]$, and $\lambda_{align} \in [0,2]$, with the zero-shot learning task.
They only affect the zero-shot learning performance marginally unless extreme values are used.
When using too large values of $\delta$ and $L$ (i.e., $\delta=0.5$ and $L=6$), the representation vector of each entity does not maintain its identity.
Meanwhile, the alignment of the cross-mode conceptual systems is not guaranteed if we use a too small value of $\lambda_{align}$ (i.e., $\lambda_{align}=0$).
Table \ref{tab:hyper}, \ref{tab:hyper2} and \ref{tab:zero-shot_param} show the empirical results for the hyperparameter search.
In Table \ref{tab:hyper2}, we made some changes to the zero-shot learning task to reduce the influence of hyperparameter $\lambda_{align}$ and highlight the impact of $\delta$ and $L$.
Specifically, we set $c_{o_iw_*}$ and $c_{o_*w_j}$ to 0 for the object-word pairs $o_i$-$w_j$ which are the evaluation targets of the zero-shot learning task, where $o_*$ and $w_*$ indicate all objects and words, respectively.

\textbf{Baseline using Spearman correlation} \quad
In the previous study \cite{roads2020learning}, the authors found that the alignment correlation (i.e., Spearman correlation) of two conceptual systems was positively correlated with the object-word mapping accuracy.
However, given the $Z$ object-word pairs, there are $Z!$ possible one-to-one mappings.
As an example, there are a total of 3,628,800 possible one-to-one mappings for 10 object-word pairs.
Thus, examining the correlation of all possible mappings cases requires excessive computational cost.
They also show that maximizing the alignment correlation does not guarantee the best mapping, especially when the number of pairs is small.
These limitations indicate difficulties in aligning conceptual systems in an unsupervised manner.
On the other hand, our method using aligned cross-modal representations overcomes these limitations.
In the zero-shot learning task, the proposed method directly measures the similarity of each object-word pair based on the aligned cross-modal representations, and uses the Hungarian algorithm to find the optimal one-to-one mapping with low computational costs.
Table \ref{tab:zero-shot_param} shows that the mapping accuracy of our model is much higher than that of the method using the alignment correlation.
Here, Spearman correlations were obtained for up to 1M potential mappings randomly selected to find the best mapping with an acceptable computational cost.

\begin{table}[h]
\caption{Mapping accuracy (mean $\pm$ std) of zero-shot learning task with various $\delta$ and $L$. The experiment was performed on 10 object-word pairs, and a value of 1.0 was used for $\lambda_{align}$. \textbf{Boldface} indicates the top three and the \textcolor{blue}{blue text} indicates the bottom three. The mean and standard deviation were drawn from the results of 20 trials.}
\label{tab:hyper}
    \begin{center}
    \begin{small}
    \begin{adjustbox}{max width=0.99\columnwidth}
        \begin{tabular}{ c c c c c c c}
        \toprule
        \multirow{2}{*}{$L$} & \multicolumn{2}{c}{$\delta = 0.1$} & \multicolumn{2}{c}{$\delta = 0.3$} & \multicolumn{2}{c}{$\delta = 0.5$} \\
        \cmidrule(r){2-3} \cmidrule(r){4-5} \cmidrule(r){6-7}
        & Perplexity & Mapping accuracy (\%) & Perplexity & Mapping accuracy (\%) & Perplexity & Mapping accuracy (\%) \\
        \midrule
        
         1 & 1.01 $\pm$ 0.01 & 69.50 $\pm$ 20.37 & 1.01 $\pm$ 0.01  & 73.50 $\pm$ 18.24 & 1.02 $\pm$ 0.02 & \textbf{75.50} $\pm$ 17.17  \\
         2 & 1.04 $\pm$ 0.06 & 68.50 $\pm$ 20.07 & 1.02 $\pm$ 0.03 & \textcolor{blue}{65.50} $\pm$ 18.83 & 1.02 $\pm$ 0.04 & 67.00 $\pm$ 22.83  \\
         3 & 1.02 $\pm$ 0.02 & \textcolor{blue}{65.00} $\pm$ 18.03 & 1.01 $\pm$ 0.01 & 74.50 $\pm$ 11.17 & 1.03 $\pm$ 0.05 & \textbf{77.50} $\pm$ 18.13  \\
         4 & 1.02 $\pm$ 0.02 & 68.00 $\pm$ 20.15 & 1.01 $\pm$ 0.01 & 74.50 $\pm$ 18.83 & 1.12 $\pm$ 0.17 & \textcolor{blue}{62.00} $\pm$ 19.64  \\
         5 & 1.01 $\pm$ 0.03 & 70.00 $\pm$ 14.83 & 1.01 $\pm$ 0.01 & 71.00 $\pm$ 19.72 & 22.63 $\pm$ 20.82 & \textbf{76.50} $\pm$ 15.90  \\
         6 & 1.01 $\pm$ 0.01 & 68.50 $\pm$ 22.20 & 1.12 $\pm$ 0.20 & 71.00 $\pm$ 17.58 & N/A & 70.00 $\pm$ 17.03  \\

        \bottomrule
        \end{tabular}
        \end{adjustbox}
    \end{small}
    \end{center}
\end{table}

\begin{table}[h]
\caption{Mapping accuracy (mean $\pm$ std) of zero-shot learning task for hyperparameter search with various $\delta$ and $L$. The experiment was performed on 10 object-word pairs, and a value of 1.0 was used for $\lambda_{align}$. In this experiment, we set $c_{o_iw_*}$ and $c_{o_*w_j}$ to 0 for the object-word pairs $o_i$-$w_j$. \textbf{Boldface} indicates the top three and the \textcolor{blue}{blue text} indicates the bottom three. The mean and standard deviation were drawn from results of 20 trials.}
\label{tab:hyper2}
    \begin{center}
    \begin{small}
    \begin{adjustbox}{max width=0.99\columnwidth}
        \begin{tabular}{ c c c c c c c}
        \toprule
        \multirow{2}{*}{$L$} & \multicolumn{2}{c}{$\delta = 0.1$} & \multicolumn{2}{c}{$\delta = 0.3$} & \multicolumn{2}{c}{$\delta = 0.5$} \\
        \cmidrule(r){2-3} \cmidrule(r){4-5} \cmidrule(r){6-7}
        & Perplexity & Mapping accuracy (\%) & Perplexity & Mapping accuracy (\%) & Perplexity & Mapping accuracy (\%) \\
        \midrule

         1 & 1.02 $\pm$ 0.02 & \textcolor{blue}{13.50} $\pm$ 11.52 & 1.02 $\pm$ 0.02 & 21.00 $\pm$ 13.75 & 1.02 $\pm$ 0.02 & 26.00 $\pm$ 16.85  \\
         2 & 1.02 $\pm$ 0.02 & 22.00 $\pm$ 12.08 & 1.01 $\pm$ 0.01 & 27.00 $\pm$ 17.91 & 1.02 $\pm$ 0.04 & 31.00 $\pm$ 15.46  \\
         3 & 1.02 $\pm$ 0.02 & \textcolor{blue}{20.00} $\pm$ 13.78 & 1.01 $\pm$ 0.01 & 31.50 $\pm$ 13.52 & 1.02 $\pm$ 0.02 & 28.50 $\pm$ 17.68  \\
         4 & 1.02 $\pm$ 0.02 & \textcolor{blue}{19.50} $\pm$ 11.17 & 1.01 $\pm$ 0.02 & \textbf{36.00} $\pm$ 22.89 & 1.11 $\pm$ 0.19 & \textbf{34.00} $\pm$ 22.45  \\
         5 & 1.02 $\pm$ 0.02 & 25.00 $\pm$ 16.88 & 1.03 $\pm$ 0.03 & \textbf{37.50} $\pm$ 18.13 & 19.52 $\pm$ 13.61 & 27.00 $\pm$ 14.18  \\
         6 & 1.01 $\pm$ 0.01 & 25.00 $\pm$ 12.04 & 1.05 $\pm$ 0.08 & 27.50 $\pm$ 18.94 & N/A & 30.00 $\pm$ 15.17   \\

        \bottomrule
        \end{tabular}
        \end{adjustbox}
    \end{small}
    \end{center}
\end{table}

\newpage
\begin{table}[h]
\caption{Mapping accuracy (mean $\pm$ std) of zero-shot learning task according to $\lambda_{align}$ value. For the hyperparameters $\delta$ and $L$, values of 0.3 and 5 were used, respectively. Perplexity (mean $\pm$ std) for self-identification was measured for all objects and words. We investigated the Spearman correlations for 1M potential mappings for 10 or 20 pairs and all 120 possible mappings for 5 pairs and reported the best mappings. For statistical significance, each experiment was repeated 30 times and correct object-word pairs were randomly selected for each trial.}
\label{tab:zero-shot_param}
    \begin{center}
    \begin{small}
    \begin{adjustbox}{max width=0.99\columnwidth}
        \begin{tabular}{c c c c c }
        \toprule
        Number of pairs & Method & $\lambda_{align}$ & Perplexity & Mapping accuracy (\%) \\
        \midrule
        
        & Random & N/A & N/A & 19.33 $\pm$ 19.56   \\
        & Spearman correlation & 1.6 & 1.03 $\pm$ 0.02 & 23.14 $\pm$ 26.68   \\
        \cmidrule(r){2-5}
        &  w/o $\mathcal{L}_{align}$ & 0.0 & 1.00 $\pm$ 0.00 & 24.00 $\pm$ 22.53   \\
        \cmidrule(r){2-5}
        & \multirow{10}{*}{Ours} &  0.2  & 1.01 $\pm$ 0.02 &  74.67 $\pm$ 27.26 \\
        & &  0.4  & 1.01 $\pm$ 0.01 &  72.00 $\pm$ 27.09 \\
        & &  0.6  & 1.02 $\pm$ 0.04 &  \textbf{87.33} $\pm$ 22.58 \\
        \multirow{2}{*}{5} & &  0.8  & 1.02 $\pm$ 0.03 &  80.67 $\pm$ 27.53 \\
        & &  1.0  & 1.03 $\pm$ 0.07 &  79.33 $\pm$ 30.39 \\
        & &  1.2  & 1.03 $\pm$ 0.04 &  69.33 $\pm$ 30.95 \\
        & &  1.4  & 1.03 $\pm$ 0.03 &  68.00 $\pm$ 28.09 \\
        & &  1.6  & 1.03 $\pm$ 0.02 &  \textbf{87.33} $\pm$ 21.96 \\
        & &  1.8  & 1.04 $\pm$ 0.03 &  81.33 $\pm$ 24.60 \\
        & &  2.0  & 1.04 $\pm$ 0.05 &  75.33 $\pm$ 28.62 \\
        \midrule
        & Random & N/A & N/A & 13.00 $\pm$ 10.88   \\
        & Spearman correlation & 1.6 & 1.03 $\pm$ 0.02 & 10.67 $\pm$ 10.15   \\
        \cmidrule(r){2-5}
        &  w/o $\mathcal{L}_{align}$ & 0.0 & 1.00 $\pm$ 0.01 & 13.33 $\pm$ 11.55   \\
        \cmidrule(r){2-5}
        & \multirow{10}{*}{Ours} & 0.2  & 1.05 $\pm$ 0.15 &  62.33 $\pm$ 22.54 \\
        & & 0.4  & 1.01 $\pm$ 0.02 &  67.67 $\pm$ 19.77 \\
        & & 0.6  & 1.01 $\pm$ 0.01 &  61.67 $\pm$ 16.63 \\
        \multirow{2}{*}{10} & & 0.8  & 1.01 $\pm$ 0.01 &  61.67 $\pm$ 21.35 \\
        & & 1.0  & 1.03 $\pm$ 0.07 &  69.67 $\pm$ 17.52 \\
        & & 1.2  & 1.03 $\pm$ 0.02 &  67.33 $\pm$ 18.18 \\
        & & 1.4  & 1.03 $\pm$ 0.03 &  \textbf{71.33} $\pm$ 19.61 \\
        & & 1.6  & 1.03 $\pm$ 0.02 &  63.33 $\pm$ 18.07 \\
        & & 1.8  & 1.04 $\pm$ 0.05 &  67.67 $\pm$ 21.92 \\
        & & 2.0  & 1.04 $\pm$ 0.03 &  65.00 $\pm$ 20.80 \\
        \midrule
        & Random & N/A & N/A & 6.17 $\pm$ 8.27   \\
        & Spearman correlation & 1.6 & 1.03 $\pm$ 0.02 & 7.17 $\pm$ 6.91   \\
        \cmidrule(r){2-5}
        & w/o $\mathcal{L}_{align}$ & 0.0 & 1.01 $\pm$ 0.03 & 5.67 $\pm$ 5.83   \\
        \cmidrule(r){2-5}
        & \multirow{10}{*}{Ours} &  0.2  & 1.04 $\pm$ 0.10 &  50.83 $\pm$ 10.75 \\
        & &  0.4  & 1.01 $\pm$ 0.01 &  52.33 $\pm$ 15.07 \\
        & &  0.6  & 1.01 $\pm$ 0.01 &  55.17 $\pm$ 11.02 \\
        \multirow{2}{*}{20} & &  0.8  & 1.01 $\pm$ 0.01 &  56.67 $\pm$ 12.27 \\
        & &  1.0  & 1.02 $\pm$ 0.03 &  51.33 $\pm$ 12.99 \\
        & &  1.2  & 1.03 $\pm$ 0.02 &  54.67 $\pm$ 13.51 \\
        & &  1.4  & 1.03 $\pm$ 0.02 &  56.67 $\pm$ 13.54 \\
        & &  1.6  & 1.03 $\pm$ 0.02 &  54.67 $\pm$ 15.70 \\
        & &  1.8  & 1.03 $\pm$ 0.02 &  \textbf{57.00} $\pm$ 11.72 \\
        & &  2.0  & 1.04 $\pm$ 0.03 &  56.83 $\pm$ 11.71 \\
        \bottomrule
        \end{tabular}
        \end{adjustbox}
    \end{small}
    \end{center}
\end{table}

\newpage
\textbf{Computational cost} \quad
Table \ref{tab:time} shows the run-time of our Python implementation on a 3.9 GHz CPU machine.
Our method has a much shorter computation time than the `Spearman correlation', and the increase in computation time is not significant as the number of pairs increases.
On the other hand, the method using `Spearman correlation' excessively increases the computation time as the number of object-word pairs increases.

\begin{table}[h]
\caption{Comparison of average run-time for zero-shot learning task according to the number of object-word pairs. Spearman correlations were investigated for 1M potential mappings for 10 or 20 pairs and for all 120 possible mappings for 5 pairs. For statistical significance, each experiment was repeated 30 times and correct object-word pairs were randomly selected for each trial.}
\label{tab:time}
    \begin{center}
    \begin{small}
    \begin{adjustbox}{max width=0.99\columnwidth}
        \begin{tabular}{c c c}
        \toprule
        Number of pairs & Method & Computation time (ms) \\
        \midrule
         \multirow{2}{*}{5} & Spearman correlation & 52.25 \\
          & Ours & 0.18 \\
        \midrule
         \multirow{2}{*}{10} & Spearman correlation & 449887.29 \\
          & Ours & 0.19 \\
        \midrule
          \multirow{2}{*}{20} & Spearman correlation & 484357.88 \\
          & Ours & 0.44 \\
        \bottomrule
        \end{tabular}
        \end{adjustbox}
    \end{small}
    \end{center}
\end{table}

\end{document}